\PassOptionsToPackage{numbers,sort&compress}{natbib}

\documentclass{article}

\usepackage[preprint]{neurips_2026}

\usepackage[utf8]{inputenc}
\usepackage[T1]{fontenc}
\usepackage{wrapfig}
\usepackage[pagebackref=false,breaklinks=true,colorlinks=True,urlcolor=citecolor,citecolor=citecolor,linkcolor=newcitecolor,bookmarks=false]{hyperref}
\usepackage{wrapfig}
\usepackage{capt-of}
\usepackage{url}
\usepackage{booktabs}
\usepackage{amsfonts}
\usepackage{nicefrac}
\usepackage{microtype}
\usepackage{xcolor}
\usepackage[table]{xcolor}
\usepackage{amsmath,amssymb,amsthm}
\usepackage{graphicx}
\usepackage{subcaption}
\usepackage{multirow}
\usepackage{makecell}
\usepackage{array}
\usepackage{enumitem}
\usepackage{mathtools}
\usepackage{float}
\usepackage{caption}
\usepackage{placeins}

\usepackage[table]{xcolor}
\definecolor{bestblue}{RGB}{170,210,250}
\definecolor{secondblue}{RGB}{225,240,255}
\newcommand{\best}[1]{\cellcolor{bestblue} #1}
\newcommand{\second}[1]{\cellcolor{secondblue} #1}

\definecolor{citecolor}{HTML}{0071bc}
\definecolor{linkc}{rgb}{0, 0.44, 0.74}
\definecolor{eqc}{rgb}{1, 0, 0}
\definecolor{newcitecolor}{rgb}{0,0.6,0}

\usepackage{booktabs}
\usepackage{caption}
\usepackage[table]{xcolor}
\usepackage{makecell}

\definecolor{lightblue}{RGB}{225,240,255}
\definecolor{midblue}{RGB}{190,220,250}
\definecolor{lightgray}{RGB}{245,245,245}

\newcommand{\bestcell}[1]{\cellcolor{midblue}\textbf{#1}}
\newcommand{\goodcell}[1]{\cellcolor{lightblue}#1}

\title{Video-Zero: Self-Evolution Video Understanding}

\author{%
  \textbf{Ruixu Zhang}$^{1,\S}$ \quad
  \textbf{Deyi Ji}$^{2,\ddagger}$ \quad
  \textbf{Lanyun Zhu}$^{3,\dagger}$ \quad
  \textbf{Xuanyi Liu}$^{4}$ \\
  \textbf{Yuxin Meng}$^{1}$ \quad
  \textbf{Ruihang Chu}$^{1,\dagger}$ \quad
  \textbf{Yujiu Yang}$^{1,\dagger}$ \\
  $^{1}$Tsinghua University \quad
  $^{2}$Tencent \quad
  $^{3}$Tongji University \quad
  $^{4}$Peking University
}

\begin{document}

\maketitle

\begingroup
\renewcommand{\thefootnote}{\fnsymbol{footnote}}
\footnotetext[2]{Corresponding authors.}
\footnotetext[3]{Project leader.}
\footnotetext[4]{Work done during an internship at Tencent.}
\endgroup

\begin{abstract}
Self-evolution offers a promising path for improving reasoning models without relying on intensive human annotation. 
However, extending this paradigm to video understanding remains underexplored and challenging: videos are long, dynamic, and redundant, while the evidence needed for reasoning is often sparse and temporally localized. 
Naively generating difficult question-answer pairs from full videos can therefore produce supervision that appears challenging but is weakly grounded, relying on static cues or language priors rather than temporal evidence. 
In this work, we argue that the key bottleneck of video self-evolution is not difficulty alone, but grounding. 
We propose \textbf{Video-Zero}, an annotation-free Questioner--Solver co-evolution framework that centers self-evolution on temporally localized evidence. 
The Questioner discovers informative evidence segments and generates evidence-grounded questions, while the Solver learns to answer and align its predictions with the supporting evidence. 
This closes an iterative loop of evidence discovery, grounded supervision, and evidence-aligned learning. 
Across 13 benchmarks spanning temporal grounding, long-video understanding, and video reasoning, Video-Zero consistently improves multiple video VLM backbones, demonstrating the effectiveness and transferability of evidence-centered self-evolution. Code and models are publicly available at \url{https://github.com/Zzz99999/Video-Zero}.
\end{abstract}

\begin{center}
    
    \includegraphics[width=\textwidth]{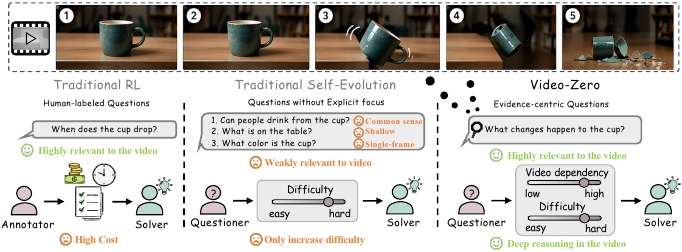}
  \captionof{figure}{
\textbf{Motivation and overview of Video-Zero.}
Traditional RL relies on costly labels; prior self-evolution often increases difficulty without explicit evidence focus.
Video-Zero grounds question generation in evidence segments and co-evolves toward video-dependent, challenging supervision.}
    \label{fig:teaser}
    \label{fig:intro}
\end{center}

\section{Introduction}

Recent advances in Vision-Language Models (VLMs) suggest that strong reasoning capabilities can emerge not only from carefully human-annotated supervision, but also from iterative self-improvement~\citep{yuan2024selfrewarding,zelikman2022star,tao2024selfevolution,chen2024spin}. 
In such self-evolution paradigms,
models generate and learn from their own supervision signals within an iterative closed loop, typically instantiated by two co-evolving roles: a \emph{Questioner} that constructs supervision from unlabeled data and a \emph{Solver} that learns from the generated questions~\citep{zhao2025absolutezero,huang2026rzero,wang2026vzero,li2026mmzero}.
Despite promising progress in text and image domains, video understanding still lacks an effective annotation-free self-evolution framework. 
This gap is particularly critical, as high-quality video supervision data is both costly and scarce~\citep{feng2025videor1,li2025videochatr1,feng2025onethinker}.

\begin{figure}[t]
\vspace{-0.6em}
    \centering
    \includegraphics[width=0.99\linewidth]{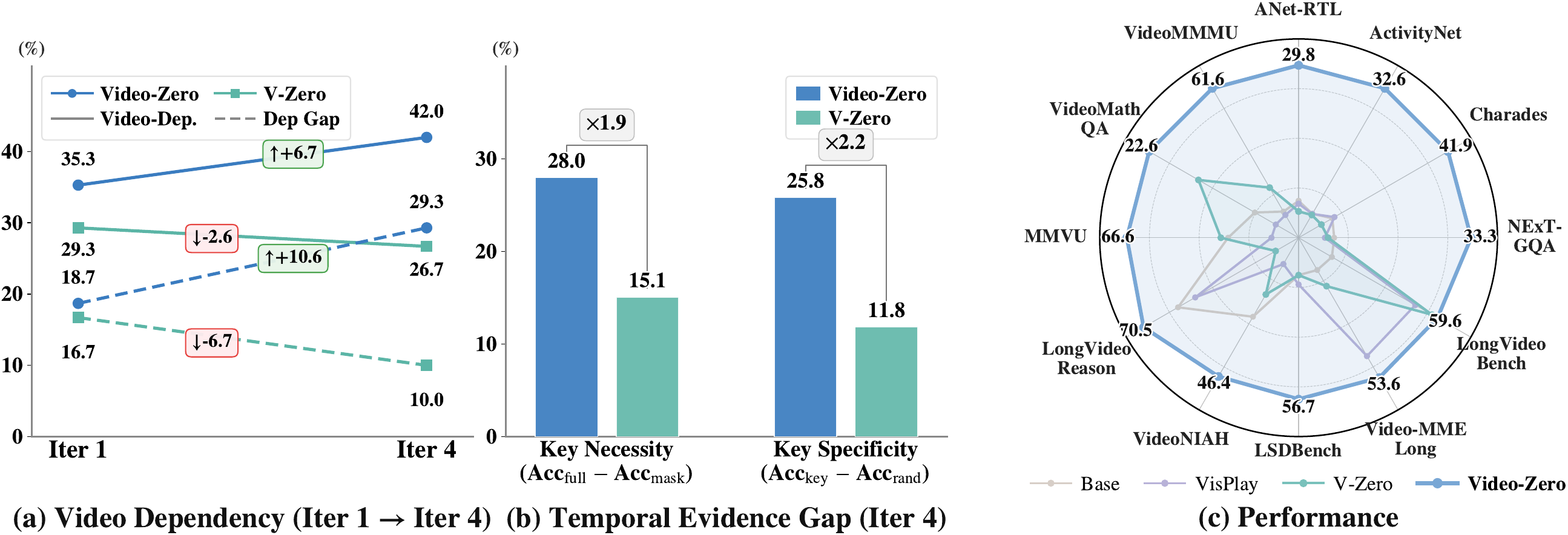}
    \vspace{-0.5em}
\caption{\textbf{Evidence-centered analysis and performance.}
Video-Zero improves (a) video dependency, (b) key-span necessity and specificity in generated questions, and (c) performance across 13 benchmarks.
Key Necessity ($\uparrow$) measures whether the key span is indispensable, while Key Specificity ($\uparrow$) measures whether it is more informative than a random span.
See Appendix~\ref{app:evidence_analysis} for details.}
    \label{fig:motivation}
    \vspace{-1em}
\end{figure}

Achieving effective self-evolution for video understanding is non-trivial.
Unlike text or images, videos are long, dynamic, and highly redundant, while task-relevant information required to construct supervision is often sparse and temporally localized~\citep{wang2025groundedvideollm}.
This characteristic exposes two fundamental limitations when naively adapting current self-evolution approaches to the video domain.
As shown in Figure~\ref{fig:teaser}, \textit{(i)}, questions are typically generated from the entire video without local focus, often resulting in queries that are weakly related to the actual video. 
Consequently, the trained model may rely only on language priors or static visual cues, rather than temporal dynamics necessary for effective video reasoning. 
This is further supported by the results in Figure~\ref{fig:motivation}, where the full-video generation method (V-Zero \citep{wang2026vzero}) produces questions with weaker dependence on the video and lower sensitivity to key temporal evidence;
\textit{(ii)} existing self-evolution loops are primarily organized around question difficulty~\citep{wang2026vzero,he2025visplay}, i.e., by optimizing the Questioner with a difficulty reward to avoid the generation of overly easy or hard samples.
Yet, training effectiveness can be affected by multiple factors beyond difficulty, particularly the relevance between the generated question and the underlying video content.
Therefore, we argue that effective self-evolution for video understanding requires a paradigm that goes beyond full-video generation and difficulty-driven optimization.

In this paper, we propose \textbf{Video-Zero}, a novel annotation-free self-evolution framework that enhances the video understanding capabilities of MLLMs directly from unlabeled videos. 
To address the above two limitations, Video-Zero features a fine-grained, temporally localized question-answer generation mechanism, along with a more comprehensive evidence-aware reward design beyond difficulty alone.
Our key design lies in an \textit{evidence-centric} manner, where temporal evidence (e.g., key events, transitions, and interactions) is explicitly inferred and leveraged for both supervision data generation and model optimization. 
Specifically, we develop a new question–answer generation mechanism conditioned on temporally localized evidence, encouraging the production of higher-quality supervision for more effective training, as evidenced by Figure~\ref{fig:motivation}(a)(b).
In addition, we introduce a comprehensive reward design that incorporates multiple factors such as question–video relevance, improving optimization for both data generation and video understanding. 
By incorporating these integrated designs, the original video MLLM is improved through a co-evolution process: an enhanced Questioner produces more informative supervision, while a stronger Solver provides more reliable feedback for subsequent evidence selection and question construction.

We validate Video-Zero across 13 benchmarks using diverse model families, including powerful Qwen3-VL family, MiMo-VL, and InternVL3.5.
Comprehensive experimental results demonstrate that Video-Zero delivers significant and consistent performance gains across temporal grounding, long-video understanding, and video reasoning—all achieved without human annotation (Figure~\ref{fig:motivation}(c)). Our main contributions are as follows:

$\bullet$ We identify a fundamental bottleneck in video self-evolution, showing that useful supervision should be anchored to temporally localized evidence rather than driven by question difficulty alone.

$\bullet$ We propose Video-Zero, an \textit{evidence-centered} co-evolution framework that unifies evidence discovery, grounded question construction, and evidence-aligned learning within a Questioner--Solver loop, requiring no human-written questions, answers, or temporal annotations.

$\bullet$ Extensive evaluations across multiple model architectures and 13 benchmarks deliver consistent and considerable improvements on diverse tasks, demonstrating the stability and scalability.

\section{Related Work}

\textbf{Self-evolution in LLMs and VLMs.}
Self-evolution reduces reliance on human annotations by letting models generate, verify, filter, and learn from their own supervision signals~\citep{wang2023selfinstruct,zelikman2022star,yuan2024selfrewarding,chen2024spin}. 
Recent reasoning-based and multimodal frameworks extend this paradigm through co-evolution, task proposal, verification, and supervision construction from unlabeled or synthetic visual inputs~\citep{zhao2025absolutezero,huang2026rzero,liu2025spice,yang2025spell,fan2026darc,wang2026vzero,he2025visplay,wang2026visionzero,he2026activezero,li2026mmzero}. 
However, they remain largely task-centric, emphasizing difficulty while leaving temporal evidence grounding under-specified. 
Video-Zero addresses this gap by making temporally localized evidence the organizing unit for question generation, solver feedback, and iterative learning.

\textbf{Video reasoning and temporal grounding.}
Recent video reasoning and post-training methods show that reinforcement learning can improve video MLLMs on video QA, temporal perception, and long-video reasoning~\citep{feng2025videor1,li2025videochatr1,wang2025timer1,ding2025videozoomer,feng2025onethinker}. 
A key trend is to encourage reasoning over relevant temporal moments through temporal rewards, evidence-aware reasoning traces, or temporally grounded supervision~\citep{feng2025videor1,li2025videochatr1,meng2025openo3video}. 
However, these pipelines typically rely on externally constructed supervision, such as curated reasoning data, cold-start traces, task-specific rewards, expert demonstrations, or temporal annotations, which are costly to scale for fine-grained video grounding~\citep{feng2025videor1,meng2025openo3video,ding2025videozoomer,feng2025onethinker}. 
In contrast, Video-Zero generates supervision directly from unlabeled videos and organizes the learning loop around temporally informative evidence.
We provide a more detailed discussion in Appendix~\ref{app:related_work}.
\section{Method}
\subsection{Overview}

\begin{figure*}[t]
    \centering
    \includegraphics[width=\textwidth]{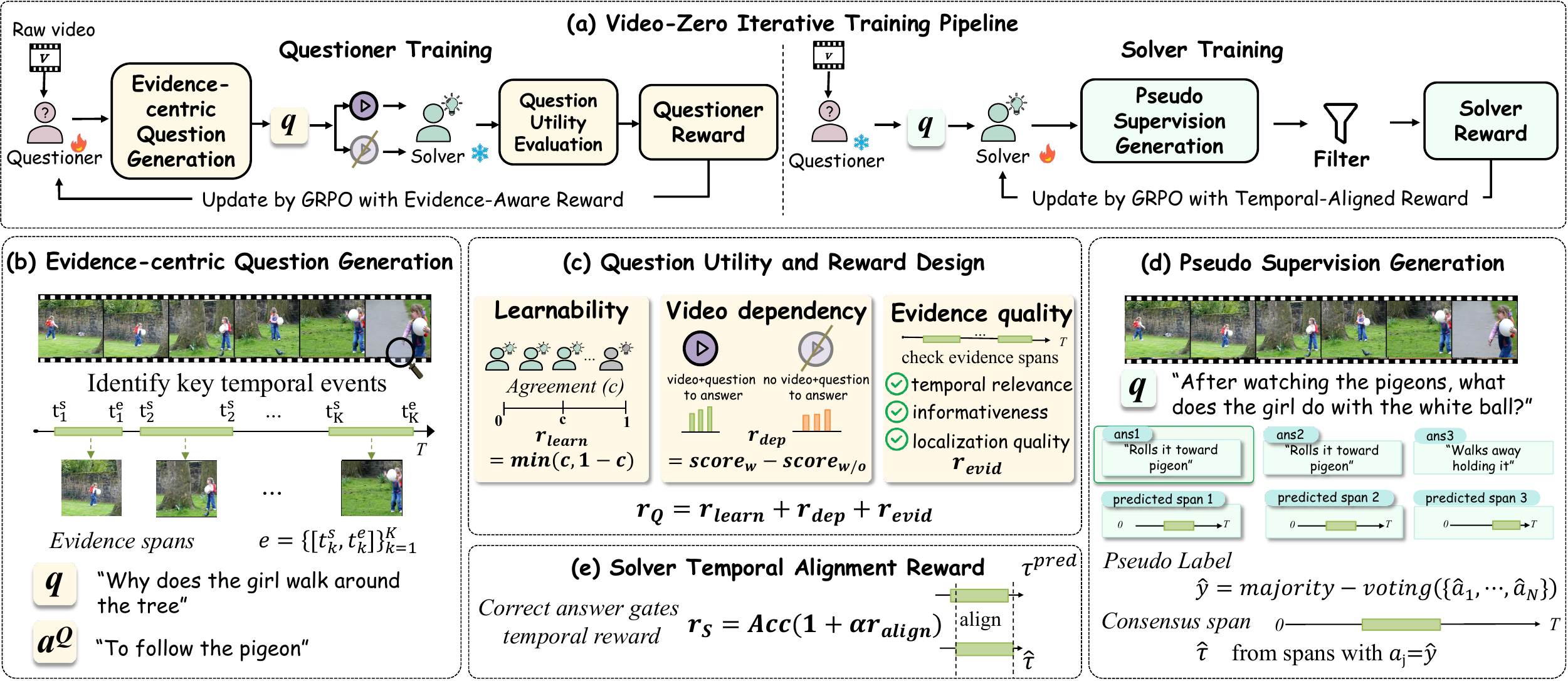}
\caption{\textbf{Overview of Video-Zero.}
(a) Video-Zero organizes video self-evolution around temporally localized evidence.
A Questioner (b) discovers key evidence spans and generates evidence-grounded questions, which (c) are scored by learnability, video dependency, and evidence quality.
A Solver then (d) learns from rollout pseudo supervision and (e) is optimized with a temporal alignment reward for evidence-aligned answer prediction, enabling closed-loop Questioner--Solver co-evolution.}
    \label{fig:framework}
\end{figure*}

We propose Video-Zero, an evidence-centered co-evolution framework for video self-evolution. Starting from unlabeled videos, Video-Zero constructs supervision signals using model-generated evidence, questions, and answers, and iteratively improves the model without human annotations.
As illustrated in Figure~\ref{fig:framework}, given an input video $v$ with duration $T$, the framework maintains a Questioner policy $\pi_Q$ and a Solver policy $\pi_S$, both initialized from the same pretrained backbone.
For each video, the Questioner first identifies one or more informative temporal evidence spans $e=\{[t_k^s,t_k^e]\}_{k=1}^{K}$, where $0 \le t_k^s < t_k^e \le T$, and then constructs an evidence-grounded supervision unit in the form of a question-answer pair $(q,a^Q)$ conditioned on $(v,e)$.
Given $(v, q)$, the Solver predicts both an answer and a temporal span that localizes the evidence supporting its reasoning. 

The key novelty of our Video-Zero lies in its evidence-centric design philosophy, reflected in three key components. \textit{(i)} The Questioner constructs questions from localized temporal evidence rather than unconstrained video-level impressions.
\textit{(ii)} The Questioner is optimized with a video-aware utility that jointly considers learnability, video dependency, and evidence quality, since difficulty alone is insufficient in the video domain.
\textit{(iii)} To prevent answer shortcuts, the Solver is trained with pseudo answer labels and temporal targets derived from rollout consensus, along with a temporal alignment reward.
These designs form a closed co-evolution loop where the Questioner provides better supervision and the Solver provides more reliable feedback for question generation.

\subsection{Questioner: Evidence-Grounded Supervision Generation}

\subsubsection{Evidence-Centric Generation Policy}

As illustrated in Figure~\ref{fig:framework}(b), the Questioner first identifies temporally informative evidence and then constructs a question conditioned on that evidence.
This turns question generation from an unconstrained process over raw videos into an evidence-guided process over localized temporal content. Formally, the Questioner implements this evidence-centric policy through the following factorization:
\begin{equation}
    \pi_Q(e,q,a^Q \mid v)
    =
    \pi_Q(e \mid v)\pi_Q(q,a^Q \mid v,e),
\end{equation}
where the temporal evidence $e$ is generated before the question $q$ and answer $a^Q$ are finalized.
The Questioner outputs structured samples containing temporal spans, evidence descriptions, question text, and answer candidates, making evidence identification explicit before question construction.

\subsubsection{Question Utility and Reward Design}
\label{sec:questioner_utility}

The evidence-centric policy only specifies how questions are generated, but does not determine which questions are useful for self-evolution.
We thus define a question utility criterion to guide Questioner optimization.
Prior self-evolution methods often rely on difficulty for this criterion; yet, in the video domain, a difficult question may still be answerable using language priors, answer patterns, or weak static cues, without requiring genuine temporal reasoning or reliance on video content.To address this, we propose a more comprehensive utility formulation with three dimensions (Fig.~\ref{fig:framework}(c)): learnability, video dependency, and evidence quality, ensuring that a useful question is \textit{(i)} learnable for the current Solver, \textit{(ii)} dependent on video content, and \textit{(iii)} grounded in high-quality temporal evidence.

\noindent\textbf{Learnability.}
We estimate learnability using the Solver's rollout consistency under video input.
For each candidate question, we query the current Solver $m$ times with access to the video, obtaining sampled answers $\{a_1,\ldots,a_m\}$.
Next, we compute the majority-vote answer $\hat{a}$ by counting occurrences of each answer in $\mathcal{A}$ among the sampled responses, define the consistency score $c\in[0,1]$ as the fraction of responses matching $\hat{a}$, and then compute the learnability reward $r_{\text{learn}}$ based on $c$. This process is formulated as follows:
\vspace{-0.5em}
\begin{equation}
    \hat{a}=\arg\max_{a\in\mathcal{A}} \sum_{j=1}^{m}\mathbf{1}[a_j=a],
    \qquad
    c=\frac{1}{m}\sum_{j=1}^{m}\mathbf{1}[a_j=\hat{a}],
    \qquad
    r_{\text{learn}}=\min(c,1-c),
\end{equation}
where $\mathbf{1}[\cdot]$ denotes the indicator function.
Consistency serves as a proxy for whether a question lies near the current Solver's learnable boundary.
Since $r_{\text{learn}}$ is maximized at intermediate consistency, it favors questions that are challenging but still learnable.
In practice, we further downweight overly easy questions with a small penalty derived from $r_{\text{learn}}$.

\textbf{Video dependency.}
Learnability assesses whether a question falls within an appropriate difficulty range for Solver's effective learning; however, it cannot distinguish genuinely video-dependent questions from those that can be solved just using non-video shortcuts.
We therefore introduce a video-dependency reward that measures reliance on visual-temporal information by comparing the Solver's agreement with the Questioner-provided answer $a^Q$ under with-video and without-video conditions.
Specifically, we sample $m$ Solver responses  $\{a_j^{\mathrm{with}}\}_{j=1}^{m}$ with video access and $m$ responses $\{a_j^{\mathrm{without}}\}_{j=1}^{m}$ without video access, and define:
\begin{equation}
\Delta_{\text{video}}
=
\frac{1}{m}\sum_{j=1}^{m}\mathbf{1}[a_j^{\mathrm{with}}=a^Q]
-
\frac{1}{m}\sum_{j=1}^{m}\mathbf{1}[a_j^{\mathrm{without}}=a^Q],
\qquad
r_{\text{dep}}=\max(\Delta_{\text{video}},0).
\end{equation}
A larger positive $\Delta_{\text{video}}$ indicates that video input is more essential for solving the question, whereas a small or negative value suggests reliance on non-video shortcuts such as language priors, answer-option biases, static cues, or generic scene knowledge.

\textbf{Evidence quality.}
Learnability and video dependency characterize question utility for the current Solver, but do not directly ensure high-quality evidence.
A question may be learnable and video-dependent while relying on evidence that is vague, overly broad, or weakly tied to key events.
We therefore introduce an evidence-quality reward computed from the structured Questioner output without external semantic labels:
$r_{\text{evid}}=s_{\text{span}}(e)+s_{\text{event}}(e)+s_{\text{temp}}(e)$, where $s_{\text{span}}$ checks span validity and localization, $s_{\text{event}}$ favors salient events or interactions, and $s_{\text{temp}}$ rewards temporal cues in the evidence description.
To avoid degenerate evidence selection, we suppress the span reward when selected spans cover most of the video without identifying distinct events.

Combining the above components, we define the final Questioner reward as
\begin{equation}
    r_Q
    =
    r_{\text{format}}
    +
    \underbrace{
    \left(
    r_{\text{learn}}
    +
    r_{\text{dep}}
    +
    r_{\text{evid}}
    \right)
    }_{U_Q}.
\end{equation}
This reward $r_Q$ consists of a format term $r_{\text{format}}$ that enforces a valid output structure, and a utility term $U_Q$ that aggregates signals indicative of high-value generated data. Note that prior to aggregation, each component is normalized to a comparable scale (implementation details are provided in the appendix). Together, these terms encourage the generated questions to be learnable for the current Solver, dependent on the video, and grounded in high-quality temporal evidence, thereby encouraging the Questioner to generate higher-quality data.

\subsection{Solver Learning with Evidence-Aligned Supervision}

The Questioner is executed and optimized based on temporal evidence. Similarly, we also construct the Solver in an evidence-centric manner to enhance its effectiveness. Specifically, the Solver is trained with evidence-aligned supervision by deriving pseudo answer labels and temporal targets from rollout consistency (Fig.~\ref{fig:framework}(d)), and is further optimized using a temporal alignment reward(Fig.~\ref{fig:framework}(e)).

\textbf{Pseudo Supervision from Rollout Consistency.}
Given a Questioner-generated question $q$, we derive both pseudo answers and temporal spans from the Solver’s rollout distribution conditioned on $(v, q)$. Specifically, for each question, the current Solver samples $n$ responses, where the $j$-th response includes an answer $f_j(v, q)$ along with a predicted temporal span $[t_s^{(j)}, t_e^{(j)}]$.
We obtain a pseudo answer label via majority voting over the answer space $\mathcal{A}$ and use the vote ratio as a confidence measure. Temporal spans from rollouts that agree with the pseudo label are then aggregated to form a consensus temporal target:
\begin{equation}
\begin{aligned}
\hat{y}
&=
\arg\max_{a\in\mathcal{A}}
\sum_{j=1}^{n}\mathbf{1}[f_j(v,q)=a],
\qquad
s
=
\frac{1}{n}
\max_{a\in\mathcal{A}}
\sum_{j=1}^{n}\mathbf{1}[f_j(v,q)=a], \\
\hat{\tau}
&=
\left[
\operatorname{median}\!\left(\{t_s^{(j)}: f_j(v,q)=\hat{y}\}\right),
\operatorname{median}\!\left(\{t_e^{(j)}: f_j(v,q)=\hat{y}\}\right)
\right].
\end{aligned}
\end{equation}
Here, $\hat{y}$ is the majority-vote pseudo label, $s$ denotes its support ratio, and $\hat{\tau}$ is the median start/end span among rollouts agreeing with $\hat{y}$.
The confidence score $s$ serves as both a reliability estimate and a curriculum signal.
We retain samples within an intermediate confidence range, filtering out those that are either trivially easy or excessively noisy.

\textbf{Temporal Alignment Reward.}
\label{sec:solver_reward}
We combine the obtained $q$, $\hat{y}$ and $\hat{\tau}$ as the pseudo supervision signal $\tilde{x}_S$. Given $\tilde{x}_S$, 
the Solver outputs a predicted answer $\hat{a}^p$ and temporal span $\tau^p=[t_s^p,t_e^p]$.
We train it with a reward combining pseudo answer correctness and temporal alignment with $\hat{\tau}=[\hat{t}_s,\hat{t}_e]$:
\begin{equation}
r_S
=
\mathbf{1}[\hat{a}^p=\hat{y}]
\left(1+\alpha r_{\text{align}}\right),
\qquad
r_{\text{align}}
=
\operatorname{IoU}(\tau^p,\hat{\tau})
\prod_{b\in\{s,e\}}
\left[
1-\frac{|t_b^p-\hat{t}_b|}{T}
\right]_+ ,
\end{equation}
where $[x]_+=\max(x,0)$ and $\alpha=0.5$ controls the temporal alignment contribution.
The multiplicative form grants alignment reward only when the predicted answer matches the pseudo label, while the boundary term encourages start- and end-time consistency.
In this way, the Solver learns not only to answer correctly, but also to localize the supporting temporal evidence.

\subsection{Iterative Co-Evolution and Optimization}

Video-Zero performs self-evolution through an alternating Questioner--Solver loop.
At iteration $t$, the current Solver $S_t$ serves as a fixed evaluator for Questioner optimization, encouraging supervision that is learnable, video-dependent, and grounded in high-quality temporal evidence.
The optimized Questioner $Q_{t+1}$ generates an evidence-grounded training set $\mathcal{D}_{t+1}$ from unlabeled videos.
For each sample, Solver rollouts derive pseudo answer labels and temporal targets, and the Solver is trained on $\mathcal{D}_{t+1}$ with evidence-aligned supervision to obtain $S_{t+1}$.
This forms the path $S_t \rightarrow Q_{t+1} \rightarrow \mathcal{D}_{t+1} \rightarrow S_{t+1}$, inducing an evidence-centered curriculum matched to the Solver's evolving capability.
As the Solver improves, it provides more reliable feedback for evidence discovery and grounded question construction, enabling the Questioner to generate stronger supervision for the next update.

Both the Questioner and Solver are optimized with Group Relative Policy Optimization (GRPO)~\citep{deepseekai2025deepseekr1}.
For each prompt $x$, we sample a group of outputs $\{o_i\}_{i=1}^{G}$, compute scalar rewards $r_i$, and form group-normalized advantages
$A_i=(r_i-\mu_g)/(\sigma_g+\epsilon)$.
The policy is then updated with the clipped GRPO objective and KL regularization.
This shared RL procedure is applied to both roles, using the Questioner reward $r_Q$ in Sec.~\ref{sec:questioner_utility} and the Solver reward $r_S$ in Sec.~\ref{sec:solver_reward}.

\section{Experiments}
We evaluate Video-Zero along two axes: whether evidence-centered self-evolution improves temporal localization, and whether this transfers to broader video understanding and reasoning.

\subsection{Experimental Setup}

\textbf{Models and baselines.}
Our main experiments use Qwen3-VL-4B-Instruct and Qwen3-VL-8B-Instruct~\citep{qwen3vl} as the shared backbones for both the Questioner and Solver.
We compare Video-Zero with the base model and two representative self-evolution baselines, VisPlay~\citep{he2025visplay} and V-Zero~\citep{wang2026vzero}, using matched budgets, iterative schedules, and evaluation protocols unless otherwise specified.
We further test cross-architecture transfer on MiMo-SFT-7B~\citep{coreteam2025mimovltechnicalreport} and InternVL3.5-4B~\citep{wang2025internvl3_5}.

\textbf{Benchmarks.}
We evaluate Video-Zero on 13 benchmarks across temporal grounding, long-video understanding, and video reasoning.
\textbf{Temporal grounding}: ANet-RTL~\citep{huang2024lita}, ActivityNet Captions~\citep{krishna2017dense}, Charades-STA~\citep{gao2017tall}, and NExT-GQA~\citep{xiao2024nextgqa}.
\textbf{Long-video understanding}: LongVideoBench~\citep{wu2024longvideobench}, MLVU~\citep{zhou2024mlvu}, Video-MME-Long~\citep{fu2025video}, LSDBench~\citep{qu2025lsdbench}, and VideoNIAH~\citep{zhao2024videoniah}.
\textbf{Video reasoning}: LongVideoReason~\citep{chen2025longvilar1}, MMVU~\citep{zhao2025mmvu}, VideoMathQA~\citep{rasheed2025videomathqa}, and VideoMMMU~\citep{hu2025video}.

\textbf{Metrics and training.}
We report mIoU and Recall for ANet-RTL, ActivityNet, and Charades, GQA at IoU 0.3/0.5 for NExT-GQA, and official scores for long-video and reasoning benchmarks.
For training, Video-Zero uses only 600 randomly sampled raw videos from Open-o3-Video~\citep{meng2025openo3video}, discarding all associated annotations.
All methods are trained for 5 iterations on the same unlabeled videos under matched budgets and evaluated with the same protocols; with details in Appendix~\ref{app:implementation}.

\subsection{Main Results}

\begin{table*}[t]
\caption{Main results on temporal grounding for Qwen3-VL-4B and Qwen3-VL-8B. The \protect\colorbox{bestblue}{best} and \protect\colorbox{secondblue}{second-best} results within each model block are highlighted in dark and light blue, respectively.}
\label{tab:main_grounding_grouped}
\centering
\scriptsize
\setlength{\tabcolsep}{3.6pt}
\renewcommand{\arraystretch}{1.08}
\resizebox{\textwidth}{!}{
\begin{tabular}{lcccccccccccccc}
\toprule
\multirow{2}{*}{\textbf{Methods}}
& \multicolumn{4}{c}{\textbf{ANet-RTL}}
& \multicolumn{4}{c}{\textbf{ActivityNet}}
& \multicolumn{4}{c}{\textbf{Charades}}
& \multicolumn{2}{c}{\textbf{NExT-GQA}} \\
\cmidrule(lr){2-5} \cmidrule(lr){6-9} \cmidrule(lr){10-13} \cmidrule(lr){14-15}
& \textbf{mIoU} & \textbf{R@0.3} & \textbf{R@0.5} & \textbf{R@0.7}
& \textbf{mIoU} & \textbf{R@0.3} & \textbf{R@0.5} & \textbf{R@0.7}
& \textbf{mIoU} & \textbf{R@0.3} & \textbf{R@0.5} & \textbf{R@0.7}
& \textbf{GQA@0.3} & \textbf{GQA@0.5} \\
\midrule
\multicolumn{15}{c}{\textit{Qwen3-VL-4B-Instruct}} \\
\midrule
Base Model
& 16.86 & 22.70 & 15.72 & 8.73
& 22.86 & 30.94 & 19.04 & 11.29
& 37.52 & 58.90 & 36.12 & 16.23
& 23.09 & 15.40 \\

VisPlay
& 16.58 & 23.14 & 13.97 & 8.73
& 22.89 & 30.98 & 19.12 & 11.31
& 37.59 & 58.82 & 36.59 & 16.42
& 23.14 & 15.33 \\

V-Zero
& 15.85 & 20.96 & 13.97 & 8.30
& 22.81 & 30.91 & 18.92 & 11.31
& 37.10 & 58.33 & 35.24 & 15.62
& 22.82 & 15.22 \\

Video-Zero (Iter 1)
& 20.86 & 28.82 & 20.09 & 8.73
& 24.23 & 32.97 & 20.22 & 12.04
& 38.75 & 60.59 & 36.91 & 17.45
& 31.66 & 20.80 \\

Video-Zero (Iter 2)
& 23.81 & 33.19 & 22.27 & 11.35
& 27.30 & 37.48 & 23.16 & 13.66
& 40.85 & 64.19 & 39.81 & 18.60
& 36.59 & 22.80 \\

Video-Zero (Iter 3)
& 27.75 & 39.30 & \best{28.82} & \best{13.97}
& 30.29 & 41.82 & 26.15 & 15.50
& \second{41.66} & 64.54 & \best{41.40} & 19.60
& \second{38.14} & \best{23.25} \\

Video-Zero (Iter 4)
& \best{29.75} & \best{44.98} & \best{28.82} & \second{13.54}
& \second{32.12} & \second{44.76} & \second{27.67} & \second{16.24}
& \best{41.85} & \best{65.11} & \second{40.81} & \best{20.03}
& \best{38.29} & \second{23.00} \\

Video-Zero (Iter 5)
& \second{29.34} & \second{42.79} & \second{27.95} & \second{13.54}
& \best{32.59} & \best{45.35} & \best{28.01} & \best{16.38}
& 41.48 & \second{64.62} & 39.92 & \second{19.95}
& 37.76 & 22.82 \\
\midrule
\rowcolor{gray!10}
\textbf{Peak $\Delta$}
& \textbf{+12.89} & \textbf{+22.28} & \textbf{+13.10} & \textbf{+5.24}
& \textbf{+9.73} & \textbf{+14.41} & \textbf{+8.97} & \textbf{+5.09}
& \textbf{+4.33} & \textbf{+6.21} & \textbf{+5.28} & \textbf{+3.80}
& \textbf{+15.20} & \textbf{+7.85} \\
\midrule

\multicolumn{15}{c}{\textit{Qwen3-VL-8B-Instruct}} \\
\midrule
Base Model
& 21.95 & 30.57 & 21.83 & 10.04
& 26.56 & 36.59 & 22.63 & 12.85
& 36.70 & 59.11 & 33.87 & \second{13.49}
& 35.93 & 21.83 \\

VisPlay
& 21.83 & 32.31 & 20.09 & 10.04
& 26.74 & 36.94 & 22.99 & 12.93
& 36.65 & 59.03 & 33.92 & 13.47
& 35.87 & 22.19 \\

V-Zero
& 22.77 & 32.31 & 20.52 & 9.17
& 26.69 & 36.79 & 23.05 & 13.00
& 36.43 & 58.63 & 33.66 & 13.41
& 35.73 & 22.11 \\

Video-Zero (Iter 1)
& \second{24.78} & 34.50 & \best{24.45} & \best{12.23}
& 26.75 & 36.87 & 23.08 & 12.99
& 36.32 & 58.58 & 32.93 & 12.82
& 37.98 & 23.55 \\

Video-Zero (Iter 2)
& 23.90 & 35.37 & \second{23.14} & \best{12.23}
& 27.79 & 38.49 & 23.94 & 13.42
& 36.37 & 58.84 & 32.77 & 12.72
& 39.04 & \second{24.31} \\

Video-Zero (Iter 3)
& 24.48 & \second{37.99} & 21.83 & \second{11.79}
& 28.99 & 40.17 & 25.20 & 14.15
& 36.84 & 59.22 & 33.60 & 13.12
& 39.29 & 24.24 \\

Video-Zero (Iter 4)
& \best{26.19} & \best{39.30} & \best{24.45} & \best{12.23}
& \best{29.87} & \best{41.77} & \best{26.23} & \best{14.66}
& \best{37.34} & \best{60.16} & \best{35.54} & \best{13.58}
& \second{39.60} & 24.26 \\

Video-Zero (Iter 5)
& 23.74 & 34.50 & 21.40 & 9.61
& \second{29.42} & \second{41.00} & \second{25.83} & \second{14.39}
& \second{37.07} & \second{60.03} & \second{34.41} & 12.93
& \best{39.96} & \best{25.10} \\
\midrule
\rowcolor{gray!10}
\textbf{Peak $\Delta$}
& \textbf{+4.24} & \textbf{+8.73} & \textbf{+2.62} & \textbf{+2.19}
& \textbf{+3.31} & \textbf{+5.18} & \textbf{+3.60} & \textbf{+1.81}
& \textbf{+0.64} & \textbf{+1.05} & \textbf{+1.67} & \textbf{+0.09}
& \textbf{+4.03} & \textbf{+3.27} \\
\bottomrule
\end{tabular}
}
\end{table*}

\textbf{Temporal grounding.}
We first examine whether organizing video self-evolution around temporal evidence improves temporally grounded video understanding.
As shown in Table~\ref{tab:main_grounding_grouped}, Video-Zero consistently improves both Qwen3-VL-4B and Qwen3-VL-8B across temporal localization benchmarks and grounded video QA, whereas prior self-evolution baselines such as VisPlay and V-Zero yield smaller or less consistent gains.
This suggests that temporal grounding benefits not merely from additional self-generated video QA supervision, but more directly from supervision explicitly organized around the video evidence that supports each answer.
The gains are especially pronounced on ANet-RTL and ActivityNet, where identifying the correct temporal segment is central to performance.
It also substantially improves NExT-GQA, indicating that the model also learns to connect answers with their supporting temporal evidence rather than only predict better timestamps.

\begin{table*}[t]
\caption{Main results on long-video understanding and video reasoning benchmarks for Qwen3-VL-4B and Qwen3-VL-8B. The \protect\colorbox{bestblue}{best} and \protect\colorbox{secondblue}{second-best} results within each model block are highlighted in dark and light blue, respectively.}
\label{tab:main_qa_vertical_grouped}
\centering
\scriptsize
\setlength{\tabcolsep}{4.2pt}
\renewcommand{\arraystretch}{1.08}
\resizebox{\textwidth}{!}{
\begin{tabular}{lccccccccc}
\toprule
\multirow{2}{*}{\textbf{Methods}}
& \multicolumn{5}{c}{\textbf{Long-video Understanding}}
& \multicolumn{4}{c}{\textbf{Video Reasoning}} \\
\cmidrule(lr){2-6}\cmidrule(lr){7-10}
& \textbf{LongVideoBench}
& \textbf{MLVU}
& \textbf{Video-MME-L}
& \textbf{LSDBench}
& \textbf{VideoNIAH}
& \textbf{LongVideoReason}
& \textbf{MMVU}
& \textbf{VideoMathQA}
& \textbf{VideoMMMU} \\
\midrule
\multicolumn{10}{c}{\textit{Qwen3-VL-4B-Instruct}} \\
\midrule
Base Model
& 57.70 & 62.60 & 51.60 & 54.68 & \second{45.78}
& 69.90 & 64.32 & 19.05 & 57.56 \\

VisPlay
& 59.20 & \best{63.50} & \second{53.20} & 54.83 & 45.26
& 69.60 & 63.36 & 18.33 & 57.44 \\

V-Zero
& \second{59.50} & \best{63.50} & 51.90 & 54.68 & 45.56
& 68.20 & 64.48 & 20.95 & \second{58.33} \\

Video-Zero (Iter 4)
& 59.10 & \second{63.40} & \best{53.60} & \second{56.52} & \best{46.37}
& \second{70.30} & \second{64.96} & \second{21.67} & \best{61.56} \\

Video-Zero (Peak)
& \best{59.60} & \second{63.40} & \best{53.60} & \best{56.67} & \best{46.37}
& \best{70.50} & \best{66.56} & \best{22.62} & \best{61.56} \\

\midrule
\multicolumn{10}{c}{\textit{Qwen3-VL-8B-Instruct}} \\
\midrule
Base Model
& 59.80 & 64.40 & 55.70 & \second{57.59} & 45.70
& 68.60 & \second{68.32} & 23.81 & 63.78 \\

VisPlay
& \second{60.50} & \best{64.90} & 54.90 & 57.44 & 45.70
& 68.10 & 67.04 & 21.43 & 62.67 \\

V-Zero
& 60.10 & 64.20 & 56.00 & 56.90 & \second{46.07}
& 69.20 & \best{68.80} & 22.14 & 63.22 \\

Video-Zero (Iter 4)
& 60.40 & 64.60 & \second{56.60} & 56.75 & \best{46.37}
& \second{70.70} & 68.00 & \second{25.48} & \second{64.22} \\

Video-Zero (Peak)
& \best{60.60} & \second{64.70} & \best{57.70} & \best{58.13} & \best{46.37}
& \best{71.00} & \second{68.32} & \best{26.19} & \best{65.33} \\
\bottomrule
\end{tabular}
}
\end{table*}

\textbf{Video QA and reasoning.}
We next examine whether Video-Zero's temporal-evidence modeling transfers beyond explicit grounding.
As shown in Table~\ref{tab:main_qa_vertical_grouped}, Video-Zero brings broad improvements across long-video understanding and video reasoning benchmarks on both backbones.
Since these tasks do not directly reward timestamps, the gains suggest that evidence-centric self-evolution improves the ability to more reliably identify, aggregate, and reason over temporally distributed video information.
Video-Zero improves most long-video benchmarks, including LongVideoBench, Video-MME-Long, LSDBench, and VideoNIAH, while remaining competitive on MLVU.
It also yields clear gains on reasoning benchmarks such as VideoMathQA and VideoMMMU, where models must combine video content with higher-level reasoning rather than recognize events.
Together, these results show that temporal evidence provides a transferable interface for video self-evolution.

\subsection{Cross-Architecture Transfer}

We further examine whether Video-Zero generalizes beyond the Qwen3-VL backbones used in our main experiments.
This setting tests whether evidence-centered self-evolution is tied to a particular model family or can serve as a more general training mechanism.
\begin{wrapfigure}[15]{r}{0.52\linewidth}
    \vspace{-0.8em}
    \centering
    \refstepcounter{table}
    \label{tab:cross_arch_qa}
    {\small \textbf{Table~\thetable:} Cross-architecture generalization results.}
    \vspace{-0.3em}
    \includegraphics[width=1.0\linewidth]{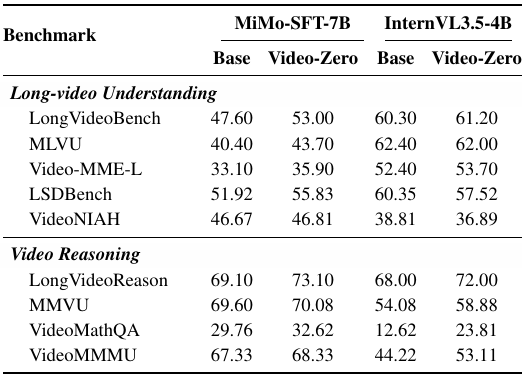}
    \vspace{-0.6em}
\end{wrapfigure}
As shown in Table~\ref{tab:cross_arch_qa}, we apply the same framework to MiMo-SFT-7B and InternVL3.5-4B without changing the pipeline.
Both backbones also improve on most benchmarks, indicating that the gains are not specific to the Qwen-VL family.
The transfer is especially clear on reasoning-intensive tasks: MiMo-SFT-7B improves across all video reasoning benchmarks and most long-video understanding benchmarks, while InternVL3.5-4B obtains large gains on MMVU, VideoMathQA, and VideoMMMU despite mixed long-video results.
These results suggest that Video-Zero transfers across architectures rather than being a backbone-specific recipe.
The consistent gains support our central design choice: organizing self-generated supervision around temporal evidence provides a model-agnostic signal for evidence identification and use over time.

\section{Analysis}

\begin{wrapfigure}[14]{r}{0.52\linewidth}
    \vspace{-1em}
    \centering
    \refstepcounter{table}
    \label{tab:ablation_main}
    {\small \textbf{Table~\thetable:} Ablation study on Qwen3-VL-4B.}
    \vspace{-0.4em}
    \includegraphics[width=1.0\linewidth]{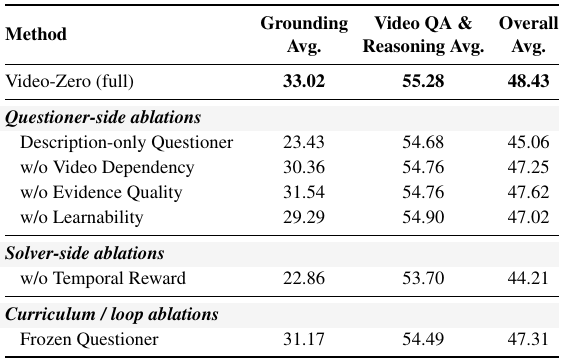}
    \vspace{-1.6em}
\end{wrapfigure}

\textbf{Ablation study.}
We ablate Questioner-side, Solver-side, and curriculum components in the evidence-centric co-evolution framework.
Table~\ref{tab:ablation_main} reports ablations on Qwen3-VL-4B, with all variants evaluated at Iteration 4.
The full model achieves the best grounding, video QA, reasoning, and overall averages, indicating that the gains come from the complete design.
A key factor is supervision generation: replacing the evidence-centric Questioner with a description-only variant causes a large drop, especially on temporal grounding.
This shows that generic video descriptions are insufficient; the Questioner must identify where the answer is supported so that generated questions remain tied to temporal evidence.
On the Questioner side, removing video dependency, evidence quality, or learnability control hurts performance, showing that useful supervision should require visual evidence, be grounded in informative temporal spans, and remain learnable for the Solver.
On the Solver side, removing the temporal reward causes the largest overall degradation, suggesting that answer correctness alone is insufficient and the Solver must align predictions with supporting evidence.
Finally, freezing the Questioner underperforms the full model, confirming that the gains arise from iterative Questioner--Solver co-evolution rather than a fixed synthetic dataset.
Together, these results support the central design of Video-Zero: evidence should guide both supervision construction and the learning signal used to absorb it.

\begin{figure*}[t]
\centering
\vspace{-0.1em}

\begin{minipage}[t]{1.0\textwidth}   
\centering

\captionof{table}{
\textbf{Questioner analysis across iterations.}
Later Questioners generate more video-dependent and challenging questions.
(a) Video dependency is reflected by the growing gap between with-video and without-video Solver inputs.
(b) Solver-checkpoint evaluation further shows that later question sets tend to best match Solvers from similar iterations, while pseudo-label accuracy decreases.
}
\label{tab:questioner_analysis}
\vspace{-0.3em}

\begin{minipage}[t]{0.36\linewidth}
\centering
\scriptsize
\setlength{\tabcolsep}{4.2pt}
\renewcommand{\arraystretch}{1.15}

\textbf{(a) Video Dependency}
\vspace{0.2em}

\begin{tabular}{lccc}
\toprule
\textbf{Q Set} & \textbf{w/ Vid} & \textbf{w/o Vid} & $\Delta_{\rm video}$ \\
\midrule
QIter1 & 56.50 & 49.00 & \goodcell{+7.50} \\
QIter2 & 50.00 & 40.50 & \goodcell{+9.50} \\
QIter3 & 52.50 & 42.50 & \goodcell{+10.00} \\
QIter4 & 54.00 & 37.50 & \goodcell{+16.50} \\
QIter5 & 49.50 & 30.50 & \bestcell{+19.00} \\
\bottomrule
\end{tabular}
\end{minipage}
\hspace{0.02\linewidth}   
\begin{minipage}[t]{0.6\linewidth}
\centering
\scriptsize
\setlength{\tabcolsep}{3.2pt}
\renewcommand{\arraystretch}{1.15}

\textbf{(b) Difficulty and Pseudo-label Reliability}
\vspace{0.2em}

\begin{tabular}{lccccccc}
\toprule
\textbf{Q Set} & \textbf{Base} & \textbf{S(1)} & \textbf{S(2)} & \textbf{S(3)} & \textbf{S(4)} & \textbf{S(5)} & \textbf{PL Acc.} \\
\midrule
QIter1 & 59.00 & \bestcell{66.00} & 61.50 & 60.00 & 61.00 & 60.50 & \bestcell{64.50} \\
QIter2 & 46.50 & 52.00 & \bestcell{54.50} & 50.00 & 50.50 & \goodcell{54.00} & \goodcell{57.50} \\
QIter3 & 41.50 & 45.50 & 45.50 & \bestcell{49.50} & 45.50 & 47.50 & 50.00 \\
QIter4 & 37.50 & 34.50 & 32.50 & 33.50 & \bestcell{44.00} & 39.00 & 40.50 \\
QIter5 & 33.00 & 35.00 & 33.00 & 30.50 & 33.50 & \bestcell{36.00} & 37.50 \\
\bottomrule
\end{tabular}

\end{minipage}

\end{minipage}

\vspace{-0.3em}
\end{figure*}

\textbf{Iterative co-evolution dynamics.}
A key question is whether the gains of Video-Zero accumulate through iterative co-evolution or mainly come from a single round of self-generated supervision.
Since temporal grounding is the capability most directly targeted by evidence-centric self-evolution, we track the average progressive grounding improvement over the base model across five iterations.

\begin{wrapfigure}[11]{r}{0.35\linewidth}
    \vspace{-0.8em}
    \centering
    \includegraphics[width=\linewidth]{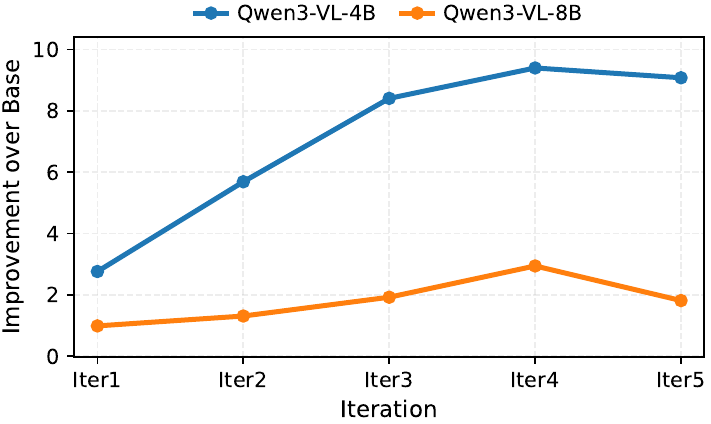}
    \vspace{-0.6em}
    \caption{Iteration improvement in temporal grounding.}
    \label{fig:iter_curve}
    \vspace{-0.8em}
\end{wrapfigure}
As shown in Figure~\ref{fig:iter_curve}, both backbones continue to improve beyond the first iteration, indicating that Video-Zero is not merely a one-shot data generation procedure.
The 4B model obtains larger iterative gains, while the 8B model improves more moderately, consistent with its stronger initialization.
Although the curves are not strictly monotonic and begin to saturate in later rounds, their overall upward trend supports iterative Questioner--Solver co-evolution as a driver of temporal evidence localization.
We analyze this behavior next by examining how the evolving Questioner changes the learnability, video dependency, and reliability of the generated supervision.

\textbf{What does the Questioner learn?} We further analyze how the Questioner evolves over iterations.
All results are computed on 200 randomly sampled examples, with pseudo-label reliability measured against Qwen3-VL-235B-A22B~\citep{qwen3vl}.
As shown in Table~\ref{tab:questioner_analysis}, later Questioners generate supervision that is more video-dependent and challenging.
The video dependency gap between with-video and without-video settings increases from +7.5 on QIter1 to +19.0 on QIter5, indicating reduced reliance on language priors or option patterns.
Meanwhile, base-Solver accuracy on the generated question sets drops from 59.0 to 33.0, showing that the Questioner progressively moves beyond easy supervision.
The best-performing Solver shifts with the Questioner iteration, forming a near-diagonal pattern from QIter1/S(1) to QIter5/S(5), suggesting that the Questioner generates questions near the contemporary Solver's learning boundary.
However, this introduces a trade-off: as questions become harder and more video-dependent, pseudo-label accuracy drops from 64.5 on QIter1 to 37.5 on QIter5.
This helps explain the saturation in Figure~\ref{fig:iter_curve}: early iterations benefit from more informative supervision, while later iterations face growing pseudo-label noise that partially offsets further gains.

\textbf{Frame-Budget Robustness.}
\label{sec:frame_budget}
\begin{figure*}[t]
    \vspace{-0.6em}
    \centering
    \includegraphics[width=\textwidth]{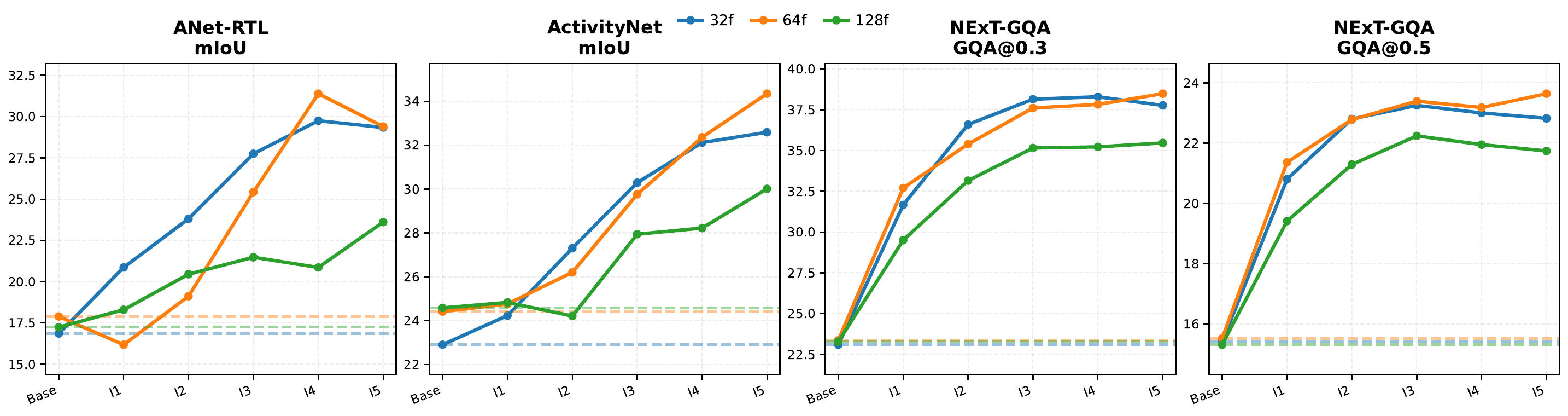}
    \vspace{-1em}
    \caption{\textbf{Frame-budget robustness.}
Video-Zero improves temporal grounding under 32/64/128 max-frame budgets at 2 FPS.
Dashed lines denote base performance, and I denotes iteration.}
    \label{fig:frame_budget_grounding}
    \vspace{-1.7em}
\end{figure*}
Figure~\ref{fig:frame_budget_grounding} examines whether the temporal grounding gains of Video-Zero depend on the visual context budget.
We vary max sampled frames among 32, 64, and 128 while keeping the sampling rate fixed at 2 FPS.
Across ANet-RTL, ActivityNet, and NExT-GQA, Video-Zero consistently improves over the corresponding base model under all three budgets, showing that the gains are not tied to a single frame budget.
The improvements are especially clear under 32f and 64f, while 128f also remains generally above its base performance.
Together with the additional robustness, scaling, seed, and qualitative analyses in Appendix~\ref{app:robustness_scaling_analysis}, these results suggest that evidence-centered self-evolution remains stable across different training and evaluation conditions.

\section{Conclusion}

In this work, we studied annotation-free self-evolution for video understanding and identified a key bottleneck: useful supervision should be grounded in temporally localized evidence rather than driven by question difficulty alone.
To address this challenge, we proposed \textbf{Video-Zero}, an evidence-centered Questioner--Solver co-evolution framework that improves video MLLMs directly from unlabeled videos.
Video-Zero discovers more informative evidence segments, constructs grounded question-answer supervision, and trains the Solver to answer questions while aligning its predictions with the supporting temporal evidence.
This iterative loop enables more reliable and transferable video self-evolution.
Across 13 benchmarks and multiple video MLLM backbones, Video-Zero consistently improves temporal grounding, long-video understanding, and video reasoning, validating the effectiveness of organizing video self-evolution around temporal evidence.

\bibliographystyle{unsrtnat}
\bibliography{main}

\appendix

\newpage
\section{Appendix Overview}
\label{app:overview}

This appendix provides more supplementary material for Video-Zero.
Appendix~\ref{app:related_work} expands the discussion of related work.
Appendix~\ref{app:implementation} provides more implementation, training, and evaluation details.
Appendix~\ref{app:more_results} reports more results, ablations, and qualitative analyses.

\section{Extended Related Work}
\label{app:related_work}

\paragraph{Self-evolution in LLMs and VLMs.}
Self-evolution aims to reduce reliance on human annotation by letting models generate, verify, filter, and learn from their own supervision signals.
Early work in language models explores self-generated instructions, rationales, rewards, and iterative policy improvement~\citep{wang2023selfinstruct,zelikman2022star,yuan2024selfrewarding,chen2024spin}.
Recent reasoning-oriented and multimodal frameworks extend this paradigm through co-evolution, task proposal, verification, and supervision construction from unlabeled or synthetic visual inputs~\citep{zhao2025absolutezero,huang2026rzero,liu2025spice,yang2025spell,fan2026darc,wang2026vzero,he2025visplay,thawakar2025evolmm,wang2026visionzero,he2026activezero,li2026mmzero,chen2025c2evocoevolvingmultimodaldata}.
These methods establish a general closed-loop recipe in which models propose tasks, evaluate their utility, and learn from retained examples.
However, existing self-evolution methods are still largely task-centric, with reward criteria focusing on difficulty.
While such criteria are effective for text and image reasoning, they do not explicitly ensure that a generated video question depends on temporally localized visual evidence.
As a result, a question may appear plausible or difficult while still being answerable from language priors, static cues, or generic scene knowledge.
Video-Zero addresses this limitation by making temporal evidence the central interface of video self-evolution: the Questioner discovers evidence spans, generates evidence-grounded questions, and the Solver learns from pseudo supervision aligned with these spans.

\paragraph{Reinforcement learning for video reasoning.}
Video MLLMs have advanced general video understanding through improved spatio-temporal modeling, instruction following, and long-context video representation~\citep{zhang2025videollama,wang2025internvideo2}.
Recent video reasoning and post-training methods further show that reinforcement learning can improve video MLLMs on video QA, temporal perception, and long-video reasoning~\citep{feng2025videor1,li2025videochatr1,wang2025timer1,ding2025videozoomer,tang2025tspo,feng2025onethinker}.
A common direction is to encourage models to reason over relevant temporal moments using temporal rewards, evidence-aware reasoning traces, or temporally grounded supervision~\citep{feng2025videor1,li2025videochatr1,wang2025timer1,guo2024trace,meng2025openo3video,video-thinker,museg2025}.
These methods demonstrate the importance of temporal objectives for video post-training.
Video-Zero is complementary to this line of work, but differs in how supervision is obtained and organized.
Most existing video RL pipelines rely on externally constructed supervision, such as curated video QA data, cold-start reasoning traces, task-specific reward labels, expert demonstrations, or temporal annotations~\citep{feng2025videor1,meng2025openo3video,ding2025videozoomer,feng2025onethinker}.
In contrast, Video-Zero starts from unlabeled videos and automatically constructs evidence-grounded supervision through a Questioner--Solver loop.
Thus, reinforcement learning is used not only to improve the Solver, but also to optimize the process by which useful video supervision is generated.

\paragraph{Temporal grounding and evidence-centric video understanding.}
Temporal grounding, moment localization, and intention forecasting require models to identify or reason about the video segments that support a language query or evolving human interactions~\citep{gao2017charades,huang2024vtimellm,wang2024grounded,zeng2024timesuite,zhang2025beyond}.
These tasks highlight a core challenge in video understanding: correct answers and predictions often depend on when events happen, how interactions unfold over time, and what temporal evidence supports them, rather than on global video context alone.
Recent video reasoning work also suggests that models may exploit shortcuts such as static frames, dataset priors, or language bias when temporal evidence is not explicitly enforced~\citep{feng2025videor1,li2025videochatr1,meng2025openo3video,wang2025timer1}.
Prior work typically uses temporal evidence as an annotation, an auxiliary training signal, or an evaluation target for a given question.
Video-Zero uses temporal evidence in a different role.
Instead of treating evidence as a downstream target, it uses evidence as the starting point for annotation-free self-evolution.
The Questioner first identifies informative temporal spans from an unlabeled video and then generates question-answer pairs grounded in those spans.
The Solver is trained to answer the question while aligning its prediction with pseudo-labeled evidence.
This design turns temporal grounding from an external supervision source into the organizing principle of video self-evolution.

\section{More Implementation Details}
\label{app:implementation}

This section summarizes the training data, iterative self-evolution pipeline, optimization settings, and evaluation protocol used in our experiments.

\subsection{Training Data and Iterative Pipeline}
\label{app:pipeline_data}

Video-Zero performs five self-evolution iterations in the main experiments.
Each iteration follows the path $S_t \rightarrow Q_{t+1} \rightarrow \mathcal{D}_{t+1} \rightarrow S_{t+1}$: the current Solver $S_t$ is used to optimize the next Questioner $Q_{t+1}$, the optimized Questioner generates evidence-grounded supervision from unlabeled videos, and the retained pseudo-labeled data are used to train the next Solver $S_{t+1}$.
Training data are regenerated at each iteration rather than reusing a fixed static dataset.

We use videos sampled from Open-o3-Video~\citep{meng2025openo3video} as the unlabeled video source.
Only raw videos are used; any associated questions, answers, or annotations are discarded.
The main experiments use only 600 randomly sampled videos, while the data-scaling study uses 1K, 2K, and 3K videos under the same self-evolution pipeline.
For each video, the Questioner generates evidence-grounded supervision units containing a temporal evidence span, a question, and an answer.
Invalid samples with missing fields, unparsable timestamps, empty answers, or invalid temporal ordering are discarded.
For the main experiments, videos are sampled at 2 FPS with a maximum frame budget of 32 frames; frame-budget scaling experiments increase the maximum budget to 64 or 128 frames.

\subsection{Pseudo Supervision and Rewards}
\label{app:reward_optimization}

For Solver training data construction, the frozen Questioner first generates candidate questions from the unlabeled videos.
The current Solver then produces $n_{\text{pl}}=10$ with-video rollouts for each question.
We obtain the pseudo answer label $\hat{y}$ by majority vote and compute its support ratio $s$:
\[
\hat{y}
=
\arg\max_{a\in\mathcal{A}}
\sum_{j=1}^{n_{\text{pl}}}\mathbf{1}[f_j(v,q)=a],
\qquad
s
=
\frac{1}{n_{\text{pl}}}
\max_{a\in\mathcal{A}}
\sum_{j=1}^{n_{\text{pl}}}\mathbf{1}[f_j(v,q)=a].
\]
Samples are retained only when $0.3 \le s \le 0.8$, filtering out questions that are either too unstable for reliable pseudo-labeling or too easy for the current Solver.
For temporal supervision, we use the Solver rollouts whose final answer agrees with $\hat{y}$ and construct the consensus temporal target by taking the median start and end times across valid predicted spans:
\[
\hat{\tau}
=
\left[
\operatorname{median}\{t_s^{(j)}: f_j(v,q)=\hat{y}\},
\operatorname{median}\{t_e^{(j)}: f_j(v,q)=\hat{y}\}
\right].
\]
If the consensus span is invalid, we fall back to the Questioner-provided evidence span when available.

The Questioner reward follows the formulation in the main paper and combines format validity, learnability, video dependency, and evidence quality.
In implementation, we normalize the utility components to comparable scales before aggregation:
\[
r_Q
=
r_{\text{format}}
+
\tilde{r}_{\text{learn}}
+
\tilde{r}_{\text{dep}}
+
\tilde{r}_{\text{evid}}.
\]
Here, $p_{\text{bdry}}$ is a small implementation-level boundary regularization term for evidence spans and is included in the evidence-quality component described in the main paper.
We set the format bonus to $0.1$.
The learnability and video-dependency terms are scaled by $0.5$ and $0.3$, respectively, before normalization, and the evidence-quality reward is clipped to a small range so that it does not dominate learnability and video dependency.
The learnability term is $r_{\text{learn}}=\min(c,1-c)$, where $c$ is the majority-answer consistency under with-video Solver rollouts.
The video-dependency term compares Solver agreement with the Questioner-provided answer under with-video and without-video conditions:
\[
\Delta_{\text{video}}
=
\frac{1}{m}\sum_{j=1}^{m}\mathbf{1}[a_j^{\mathrm{with}}=a^Q]
-
\frac{1}{m}\sum_{j=1}^{m}\mathbf{1}[a_j^{\mathrm{without}}=a^Q],
\qquad
r_{\text{dep}}=\max(\Delta_{\text{video}},0).
\]

For Solver optimization, we use the evidence-aligned reward:
\[
r_S
=
\mathbf{1}[\hat{a}^p=\hat{y}]
\left(1+\alpha r_{\text{align}}\right),
\qquad
r_{\text{align}}
=
\operatorname{IoU}(\tau^p,\hat{\tau})
\prod_{b\in\{s,e\}}
\left[
1-\frac{|t_b^p-\hat{t}_b|}{T}
\right]_+ .
\]
We set $\alpha=0.5$.
Invalid or unparsable Solver outputs receive zero answer correctness and therefore no temporal alignment reward.

\subsection{Optimization Settings}
\label{app:optimization_settings}

Both the Questioner and Solver are optimized with GRPO.
For both roles, we use a learning rate of $1\times10^{-6}$, one training epoch per iteration, and a KL coefficient of $0.01$.
The Questioner uses four sampled rollouts per prompt during optimization.
The Solver uses five sampled rollouts per prompt during GRPO training and $n_{\text{pl}}=10$ rollouts per generated question for pseudo-label construction.
All training is conducted with bfloat16 precision and vLLM-based rollout generation on 8 NVIDIA H20 GPUs.
Each self-evolution iteration takes approximately 7 hours.
For fair comparison, Video-Zero and competing self-evolution baselines use the same hardware configuration, rollout budget, number of iterations, and optimization schedule unless otherwise specified.

\begin{table}[t]
\centering
\caption{Key optimization settings used in Video-Zero.}
\label{tab:optimization_settings}
\small
\setlength{\tabcolsep}{5pt}
\renewcommand{\arraystretch}{1.1}
\begin{tabular}{lcc}
\toprule
\textbf{Setting} & \textbf{Questioner} & \textbf{Solver} \\
\midrule
Optimization algorithm & GRPO & GRPO \\
Learning rate & $1\times10^{-6}$ & $1\times10^{-6}$ \\
Epochs per iteration & 1 & 1 \\
KL coefficient & 0.01 & 0.01 \\
Rollouts per prompt for GRPO & 4 & 5 \\
Rollouts for pseudo-labeling & -- & 10 \\
Max response length & 4096 & 2048 \\
Training precision & bfloat16 & bfloat16 \\
Rollout engine & vLLM & vLLM \\
\bottomrule
\end{tabular}
\end{table}

\subsection{Evaluation Protocol}
\label{app:evaluation}

We evaluate the 13 benchmarks using three types of evaluation protocols.
For ActivityNet~\citep{krishna2017dense}, ANet-RTL~\citep{huang2024lita}, Charades-STA~\citep{gao2017tall}, LongVideoReason~\citep{chen2025longvilar1}, MMVU~\citep{zhao2025mmvu}, VideoMathQA~\citep{rasheed2025videomathqa}, and VideoMMMU~\citep{hu2025video}, we follow the evaluation setting of OneThinker\citep{feng2025onethinker}.
For LongVideoBench~\citep{wu2024longvideobench}, MLVU~\citep{zhou2024mlvu}, and Video-MME-Long~\citep{fu2025video}, we use VLMEvalKit~\citep{duan2024vlmevalkit}.
For NExT-GQA~\citep{xiao2024nextgqa}, LSDBench~\citep{qu2025lsdbench}, and VideoNIAH~\citep{zhao2024videoniah}, we follow the official evaluation protocols of the original benchmarks.
All checkpoints and baselines use the same prompt template, frame budget, decoding setting, scoring function, and output parser within each benchmark.

For temporal grounding benchmarks, the model predicts a temporal span $p=[p_s,p_e]$ for each query.
We compute temporal Intersection-over-Union with the ground-truth span $g=[g_s,g_e]$:
\[
\mathrm{tIoU}(p,g)
=
\frac{
\max(0,\min(p_e,g_e)-\max(p_s,g_s))
}{
\max(p_e,g_e)-\min(p_s,g_s)
}.
\]
We report mean temporal IoU and recall at tIoU thresholds 0.3, 0.5, and 0.7.
For NExT-GQA, a prediction is counted as correct only when both the answer is correct and the predicted temporal span reaches the specified tIoU threshold.
Unparsable answers, missing spans, or invalid temporal spans are scored as incorrect.

We use deterministic decoding for the OneThinker-style and official benchmark evaluations, and follow the default VLMEvalKit configuration for LongVideoBench, MLVU, and Video-MME-Long.
All decoding settings are fixed per benchmark and shared by all compared methods.
Table~\ref{tab:eval_protocol} summarizes the split, number of evaluated samples, metric, and evaluation protocol for each benchmark.
For VLMEvalKit benchmarks, LongVideoBench contains 1{,}337 validation questions, MLVU contains 2{,}592 questions, and Video-MME-Long is obtained by evaluating the full Video-MME benchmark and reporting the 900-question long-video subset.

\begin{table*}[t]
\centering
\caption{Evaluation protocol for the 13 benchmarks reported in the main paper.}
\label{tab:eval_protocol}
\small
\setlength{\tabcolsep}{5.2pt}
\renewcommand{\arraystretch}{1.10}
\resizebox{0.82\textwidth}{!}{
\begin{tabular}{llrl}
\toprule
\textbf{Benchmark} & \textbf{Split / Subset} & \textbf{\#Samples} & \textbf{Metric} \\
\midrule
\multicolumn{4}{l}{\textit{Temporal Grounding}} \\
\midrule
ActivityNet  & full & 17{,}031 & mIoU, R@0.3/0.5/0.7 \\
ANet-RTL     & full & 228      & mIoU, R@0.3/0.5/0.7 \\
Charades-STA & full & 3{,}720  & mIoU, R@0.3/0.5/0.7 \\
NExT-GQA     & test & 5{,}554  & GQA@0.3, GQA@0.5    \\
\midrule
\multicolumn{4}{l}{\textit{Long-video Understanding}} \\
\midrule
LongVideoBench & val         & 1{,}337          & Accuracy          \\
MLVU           & test        & 2{,}592          & Accuracy          \\
Video-MME-Long & long subset & 900              & Accuracy          \\
LSDBench       & test        & 1{,}304          & Accuracy          \\
VideoNIAH      & main-4try   & 1{,}350$\times$4 & 4-try strict acc. \\
\midrule
\multicolumn{4}{l}{\textit{Video Reasoning}} \\
\midrule
LongVideoReason & full & 1{,}000 & Accuracy \\
MMVU            & full & 625     & Accuracy \\
VideoMathQA     & full & 222     & Accuracy \\
VideoMMMU       & full & 900     & Accuracy \\
\bottomrule
\end{tabular}
}
\end{table*}

\section{More Experimental Results}
\label{app:more_results}

This section provides additional experimental results that complement the main paper.
We first report full iterative QA and reasoning results on Qwen3-VL-4B-Instruct and Qwen3-VL-8B-Instruct~\citep{qwen3vl}, including all intermediate checkpoints for Video-Zero and the competing self-evolution baselines VisPlay~\citep{he2025visplay} and V-Zero~\citep{wang2026vzero}.
We then present cross-architecture results on MiMo-SFT-7B~\citep{coreteam2025mimovltechnicalreport} and InternVL-4B~\citep{wang2025internvl3_5} to examine whether the evidence-centered self-evolution procedure transfers beyond the primary Qwen3-VL backbones.
Finally, we provide full ablation results and qualitative examples to further analyze the contribution of each component and the evolution of Questioner-generated supervision.

\subsection{Full Iterative Temporal Grounding Results}
\label{app:full_grounding_results}

Table~\ref{tab:grounding_full_all_iters} reports the full temporal grounding trajectories on Qwen3-VL-4B and Qwen3-VL-8B.
While Table~\ref{tab:main_grounding_grouped} in the main paper provides a compact comparison using the Iter-4 checkpoints of VisPlay and V-Zero, this appendix table includes all intermediate checkpoints for Video-Zero and the self-evolution baselines across iterations.

\begin{table*}[t]
\caption{Full iterative temporal grounding results on Qwen3-VL-4B and Qwen3-VL-8B.
We report all intermediate checkpoints for Video-Zero, VisPlay, and V-Zero.
The best and second-best results within each model block are highlighted in dark and light blue, respectively.}
\label{tab:grounding_full_all_iters}
\centering
\tiny
\setlength{\tabcolsep}{2.6pt}
\renewcommand{\arraystretch}{1.05}
\resizebox{\textwidth}{!}{
\begin{tabular}{lcccccccccccccc}
\toprule
\multirow{2}{*}{\textbf{Methods}}
& \multicolumn{4}{c}{\textbf{ANet-RTL}}
& \multicolumn{4}{c}{\textbf{ActivityNet}}
& \multicolumn{4}{c}{\textbf{Charades}}
& \multicolumn{2}{c}{\textbf{NExT-GQA}} \\
\cmidrule(lr){2-5} \cmidrule(lr){6-9} \cmidrule(lr){10-13} \cmidrule(lr){14-15}
& \textbf{mIoU} & \textbf{R@0.3} & \textbf{R@0.5} & \textbf{R@0.7}
& \textbf{mIoU} & \textbf{R@0.3} & \textbf{R@0.5} & \textbf{R@0.7}
& \textbf{mIoU} & \textbf{R@0.3} & \textbf{R@0.5} & \textbf{R@0.7}
& \textbf{GQA@0.3} & \textbf{GQA@0.5} \\
\midrule
\multicolumn{15}{c}{\textit{Qwen3-VL-4B-Instruct}} \\
\midrule
Base Model
& 16.86 & 22.70 & 15.72 & 8.73
& 22.86 & 30.94 & 19.04 & 11.29
& 37.52 & 58.90 & 36.12 & 16.23
& 23.09 & 15.40 \\
\midrule
VisPlay (Iter 1)
& 16.29 & 24.02 & 14.85 & 7.42
& 22.93 & 31.19 & 19.08 & 11.17
& 37.46 & 58.98 & 36.18 & 16.51
& 23.30 & 15.49 \\
VisPlay (Iter 2)
& 16.44 & 22.27 & 14.85 & 8.30
& 22.76 & 30.84 & 18.82 & 11.12
& 37.49 & 58.66 & 36.18 & 16.37
& 23.36 & 15.72 \\
VisPlay (Iter 3)
& 15.34 & 21.83 & 13.54 & 7.42
& 22.83 & 31.05 & 19.14 & 11.19
& 37.20 & 58.04 & 35.91 & 16.56
& 23.34 & 15.58 \\
VisPlay (Iter 4)
& 16.58 & 23.14 & 13.97 & 8.73
& 22.89 & 30.98 & 19.12 & 11.31
& 37.59 & 58.82 & 36.59 & 16.42
& 23.14 & 15.33 \\
VisPlay (Iter 5)
& 17.23 & 23.58 & 14.85 & 8.30
& 22.81 & 30.99 & 18.97 & 11.20
& 37.35 & 58.82 & 36.13 & 16.26
& 23.25 & 15.56 \\
\midrule
V-Zero (Iter 1)
& 16.43 & 23.14 & 14.41 & 8.73
& 23.07 & 31.24 & 19.17 & 11.40
& 37.46 & 58.66 & 36.13 & 16.69
& 23.07 & 15.52 \\
V-Zero (Iter 2)
& 16.19 & 22.27 & 13.97 & 8.73
& 22.90 & 30.88 & 19.05 & 11.45
& 37.53 & 59.22 & 36.26 & 15.94
& 23.41 & 15.61 \\
V-Zero (Iter 3)
& 15.88 & 22.27 & 13.54 & 7.86
& 22.63 & 30.55 & 18.85 & 11.28
& 37.12 & 57.98 & 35.51 & 16.53
& 23.50 & 15.61 \\
V-Zero (Iter 4)
& 15.85 & 20.96 & 13.97 & 8.30
& 22.81 & 30.91 & 18.92 & 11.31
& 37.10 & 58.33 & 35.24 & 15.62
& 22.82 & 15.22 \\
V-Zero (Iter 5)
& 16.65 & 22.71 & 15.72 & 8.73
& 23.12 & 31.50 & 19.22 & 11.41
& 37.55 & 58.84 & 36.59 & 16.29
& 22.64 & 15.05 \\
\midrule
Video-Zero (Iter 1)
& 20.86 & 28.82 & 20.09 & 8.73
& 24.23 & 32.97 & 20.22 & 12.04
& 38.75 & 60.59 & 36.91 & 17.45
& 31.66 & 20.80 \\
Video-Zero (Iter 2)
& 23.81 & 33.19 & 22.27 & 11.35
& 27.30 & 37.48 & 23.16 & 13.66
& 40.85 & 64.19 & 39.81 & 18.60
& 36.59 & 22.80 \\
Video-Zero (Iter 3)
& 27.75 & 39.30 & \best{28.82} & \best{13.97}
& 30.29 & 41.82 & 26.15 & 15.50
& \second{41.66} & 64.54 & \best{41.40} & 19.60
& \second{38.14} & \best{23.25} \\
Video-Zero (Iter 4)
& \best{29.75} & \best{44.98} & \best{28.82} & \second{13.54}
& \second{32.12} & \second{44.76} & \second{27.67} & \second{16.24}
& \best{41.85} & \best{65.11} & \second{40.81} & \best{20.03}
& \best{38.29} & \second{23.00} \\
Video-Zero (Iter 5)
& \second{29.34} & \second{42.79} & \second{27.95} & \second{13.54}
& \best{32.59} & \best{45.35} & \best{28.01} & \best{16.38}
& 41.48 & \second{64.62} & 39.92 & \second{19.95}
& 37.76 & 22.82 \\
\midrule

\multicolumn{15}{c}{\textit{Qwen3-VL-8B-Instruct}} \\
\midrule
Base Model
& 21.95 & 30.57 & 21.83 & 10.04
& 26.56 & 36.59 & 22.63 & 12.85
& 36.70 & 59.11 & 33.87 & 13.49
& 35.93 & 21.83 \\
\midrule
VisPlay (Iter 1)
& 22.17 & 31.00 & 21.40 & 11.35
& 26.66 & 36.68 & 23.13 & 12.88
& 36.66 & 58.92 & 33.71 & 13.41
& 36.00 & 22.04 \\
VisPlay (Iter 2)
& 22.35 & 31.88 & 20.09 & 10.92
& 26.85 & 37.28 & 23.11 & 12.94
& 36.67 & 58.60 & 34.35 & 13.44
& 35.87 & 22.31 \\
VisPlay (Iter 3)
& 21.21 & 31.00 & 18.34 & 9.17
& 26.57 & 36.85 & 22.99 & 12.73
& 36.58 & 58.76 & 33.44 & 13.52
& 36.16 & 22.22 \\
VisPlay (Iter 4)
& 21.83 & 32.31 & 20.09 & 10.04
& 26.74 & 36.94 & 22.99 & 12.93
& 36.65 & 59.03 & 33.92 & 13.47
& 35.87 & 22.19 \\
VisPlay (Iter 5)
& 21.95 & 32.31 & 19.21 & 9.61
& 26.75 & 36.94 & 23.18 & 13.03
& 36.70 & 59.11 & 34.01 & 13.33
& 36.03 & 22.20 \\
\midrule
V-Zero (Iter 1)
& 23.36 & 34.93 & 21.83 & 10.04
& 26.89 & 37.16 & 23.29 & 13.11
& 36.43 & 58.44 & 33.33 & 13.52
& 35.78 & 22.19 \\
V-Zero (Iter 2)
& 23.05 & 31.88 & 19.21 & 10.04
& 26.76 & 36.88 & 23.01 & 13.09
& 36.66 & 59.11 & 33.52 & 13.31
& 35.51 & 22.19 \\
V-Zero (Iter 3)
& 24.70 & 34.50 & 22.71 & \best{12.23}
& 26.86 & 37.21 & 23.23 & 12.95
& 36.66 & 59.19 & 33.76 & 13.49
& 35.82 & 22.01 \\
V-Zero (Iter 4)
& 22.77 & 32.31 & 20.52 & 9.17
& 26.69 & 36.79 & 23.05 & 13.00
& 36.43 & 58.63 & 33.66 & 13.41
& 35.73 & 22.11 \\
V-Zero (Iter 5)
& 22.87 & 32.75 & 20.09 & 10.48
& 26.83 & 36.83 & 23.25 & 13.12
& 36.42 & 58.23 & 33.41 & \best{13.66}
& 35.62 & 21.95 \\
\midrule
Video-Zero (Iter 1)
& \second{24.78} & 34.50 & \best{24.45} & \best{12.23}
& 26.75 & 36.87 & 23.08 & 12.99
& 36.32 & 58.58 & 32.93 & 12.82
& 37.98 & 23.55 \\
Video-Zero (Iter 2)
& 23.90 & 35.37 & \second{23.14} & \best{12.23}
& 27.79 & 38.49 & 23.94 & 13.42
& 36.37 & 58.84 & 32.77 & 12.72
& 39.04 & \second{24.31} \\
Video-Zero (Iter 3)
& 24.48 & \second{37.99} & 21.83 & \second{11.79}
& 28.99 & 40.17 & 25.20 & 14.15
& 36.84 & 59.22 & 33.60 & 13.12
& 39.29 & 24.24 \\
Video-Zero (Iter 4)
& \best{26.19} & \best{39.30} & \second{23.14} & \best{12.23}
& \best{29.87} & \best{41.77} & \best{26.23} & \best{14.66}
& \best{37.34} & \best{60.16} & \best{35.54} & \second{13.58}
& \second{39.60} & 24.26 \\
Video-Zero (Iter 5)
& 23.74 & 34.50 & 21.40 & 9.61
& \second{29.42} & \second{41.00} & \second{25.83} & \second{14.39}
& \second{37.07} & \second{60.03} & \second{34.41} & 12.93
& \best{39.96} & \best{25.10} \\
\bottomrule
\end{tabular}
}
\end{table*}

Across both model scales, Video-Zero consistently improves temporal grounding over the base model.
The gains are especially pronounced on the 4B backbone, where performance increases steadily across most metrics from early to later iterations.
For the stronger 8B backbone, improvements are smaller but remain consistent, with the best checkpoints typically appearing in later iterations.
Compared with competing self-evolution baselines, Video-Zero shows a clearer upward trajectory, supporting the effectiveness of evidence-centered supervision generation and temporal alignment for localization-oriented video understanding.

\subsection{Full Iterative Video QA and Reasoning Results}
\label{app:full_qa_results}

Table~\ref{tab:qa_full_all_iters} reports the full iterative results on long-video understanding and video reasoning benchmarks for Qwen3-VL-4B and Qwen3-VL-8B~\citep{qwen3vl}.
While Table~\ref{tab:main_qa_vertical_grouped} in the main paper provides a compact summary using the Iter-4 checkpoints of VisPlay, V-Zero, and Video-Zero, together with the peak Video-Zero checkpoint across iterations, this appendix table includes all intermediate checkpoints for all self-evolution methods.
These results complement the temporal grounding analysis by evaluating whether evidence-centered self-evolution transfers to broader video understanding and reasoning tasks.

\begin{table*}[t]
\caption{Full iterative video QA and reasoning results on Qwen3-VL-4B and Qwen3-VL-8B.
We report all intermediate checkpoints for Video-Zero, VisPlay, and V-Zero.
For clarity, dark and light blue highlight the best and second-best Video-Zero checkpoints within each model block, respectively.}
\label{tab:qa_full_all_iters}
\centering
\tiny
\setlength{\tabcolsep}{3.0pt}
\renewcommand{\arraystretch}{1.05}
\resizebox{\textwidth}{!}{
\begin{tabular}{lccccccccc}
\toprule
\multirow{2}{*}{\textbf{Methods}}
& \multicolumn{5}{c}{\textbf{Long-video Understanding}}
& \multicolumn{4}{c}{\textbf{Video Reasoning}} \\
\cmidrule(lr){2-6}\cmidrule(lr){7-10}
& \textbf{LongVideoBench}
& \textbf{MLVU}
& \textbf{Video-MME-L}
& \textbf{LSDBench}
& \textbf{VideoNIAH}
& \textbf{LongVideoReason}
& \textbf{MMVU}
& \textbf{VideoMathQA}
& \textbf{VideoMMMU} \\
\midrule
\multicolumn{10}{c}{\textit{Qwen3-VL-4B-Instruct}} \\
\midrule
Base Model
& 57.70 & 62.60 & 51.60 & 54.68 & 45.78
& 69.90 & 64.32 & 19.05 & 57.56 \\
\midrule
VisPlay (Iter 1)
& 58.90 & 62.70 & 51.80 & 54.75 & 45.78
& 69.50 & 64.32 & 18.81 & 57.89 \\
VisPlay (Iter 2)
& 59.60 & 63.50 & 51.10 & 54.91 & 45.33
& 70.30 & 63.52 & 20.71 & 56.78 \\
VisPlay (Iter 3)
& 58.00 & 62.70 & 51.80 & 54.83 & 45.56
& 69.60 & 63.36 & 20.71 & 57.22 \\
VisPlay (Iter 4)
& 59.20 & 63.50 & 53.20 & 54.83 & 45.26
& 69.60 & 63.36 & 18.33 & 57.44 \\
VisPlay (Iter 5)
& 59.30 & 62.50 & 50.40 & 54.52 & 45.33
& 69.30 & 64.80 & 19.52 & 57.11 \\
\midrule
V-Zero (Iter 1)
& 58.60 & 62.80 & 51.90 & 54.60 & 45.63
& 69.90 & 64.16 & 20.48 & 57.67 \\
V-Zero (Iter 2)
& 59.20 & 63.10 & 52.10 & 54.75 & 45.63
& 70.00 & 64.80 & 20.48 & 56.33 \\
V-Zero (Iter 3)
& 59.60 & 63.40 & 52.80 & 54.60 & 45.63
& 68.40 & 64.32 & 19.52 & 57.67 \\
V-Zero (Iter 4)
& 59.50 & 63.50 & 51.90 & 54.68 & 45.56
& 68.20 & 64.48 & 20.95 & 58.33 \\
V-Zero (Iter 5)
& 59.20 & 62.80 & 52.00 & 54.83 & 45.48
& 69.60 & 64.16 & 20.48 & 57.89 \\
\midrule
Video-Zero (Iter 1)
& 58.20 & 62.90 & 51.80 & 54.98 & 45.70
& 68.00 & 64.32 & 19.76 & 57.11 \\
Video-Zero (Iter 2)
& \second{59.40} & 63.00 & 51.30 & 55.21 & 45.70
& 69.90 & \best{66.56} & 19.05 & 58.89 \\
Video-Zero (Iter 3)
& 59.10 & 62.80 & \second{52.80} & 55.29 & 45.78
& 69.90 & 63.20 & 21.19 & 59.00 \\
Video-Zero (Iter 4)
& 59.10 & \best{63.40} & \best{53.60} & \second{56.52} & \best{46.37}
& \second{70.30} & 64.96 & \second{21.67} & \best{61.56} \\
Video-Zero (Iter 5)
& \best{59.60} & \second{62.90} & 52.20 & \best{56.67} & \second{46.07}
& \best{70.50} & \second{65.60} & \best{22.62} & \second{60.44} \\
\midrule

\multicolumn{10}{c}{\textit{Qwen3-VL-8B-Instruct}} \\
\midrule
Base Model
& 59.80 & 64.40 & 55.70 & 57.59 & 45.70
& 68.60 & 68.32 & 23.81 & 63.78 \\
\midrule
VisPlay (Iter 1)
& 59.80 & 64.10 & 55.90 & 57.36 & 45.85
& 68.50 & 68.16 & 24.76 & 64.11 \\
VisPlay (Iter 2)
& 60.70 & 64.40 & 55.40 & 57.44 & 45.63
& 69.40 & 67.68 & 21.67 & 62.22 \\
VisPlay (Iter 3)
& 60.00 & 64.00 & 56.60 & 57.21 & 45.70
& 68.40 & 67.68 & 23.33 & 63.67 \\
VisPlay (Iter 4)
& 60.50 & 64.90 & 54.90 & 57.44 & 45.70
& 68.10 & 67.04 & 21.43 & 62.67 \\
VisPlay (Iter 5)
& 60.40 & 64.40 & 56.40 & 57.59 & 45.85
& 68.20 & 67.36 & 22.38 & 64.22 \\
\midrule
V-Zero (Iter 1)
& 60.40 & 64.60 & 54.30 & 57.06 & 45.63
& 68.40 & 67.52 & 24.05 & 64.33 \\
V-Zero (Iter 2)
& 59.80 & 63.90 & 55.00 & 57.29 & 45.70
& 69.30 & 68.48 & 22.86 & 64.22 \\
V-Zero (Iter 3)
& 60.60 & 64.10 & 55.90 & 57.21 & 45.78
& 68.10 & 67.36 & 22.38 & 63.67 \\
V-Zero (Iter 4)
& 60.10 & 64.20 & 56.00 & 56.90 & 46.07
& 69.20 & 68.80 & 22.14 & 63.22 \\
V-Zero (Iter 5)
& 60.30 & 64.60 & 56.00 & 56.90 & 45.70
& 69.90 & 68.96 & 24.52 & 62.22 \\
\midrule
Video-Zero (Iter 1)
& \best{60.60} & 64.40 & 55.90 & \best{58.13} & 45.85
& 69.70 & 67.20 & \best{26.19} & 63.78 \\
Video-Zero (Iter 2)
& 60.20 & \best{64.70} & \second{57.00} & \second{57.29} & \second{46.07}
& 68.40 & \best{68.32} & 23.33 & 63.11 \\
Video-Zero (Iter 3)
& \second{60.40} & 63.20 & \best{57.70} & \second{57.29} & 45.93
& 69.20 & 67.36 & 25.24 & 63.56 \\
Video-Zero (Iter 4)
& \second{60.40} & \second{64.60} & 56.60 & 56.75 & \best{46.37}
& \second{70.70} & 68.00 & 25.48 & \second{64.22} \\
Video-Zero (Iter 5)
& 60.20 & 63.80 & 56.70 & 56.06 & \second{46.15}
& \best{71.00} & \second{68.16} & \second{25.95} & \best{65.33} \\
\bottomrule
\end{tabular}
}
\end{table*}

Video-Zero generally improves over the base model across both model scales, with gains on benchmarks requiring long-context understanding, temporal evidence retrieval, or multi-step reasoning.
Different benchmarks peak at different iterations, suggesting that the co-evolution process provides complementary benefits across task types rather than a single uniformly optimal checkpoint.
Together with the temporal grounding results, these trajectories show that evidence-centered co-evolution provides useful training signals beyond temporal localization alone.

\subsection{Cross-Architecture Results}
\label{app:cross_architecture_results}

Table~\ref{tab:cross_arch_qa_side_by_side} reports the full cross-architecture results on MiMo-SFT-7B~\citep{coreteam2025mimovltechnicalreport} and InternVL3.5-4B~\citep{wang2025internvl3_5}.
While Table~\ref{tab:cross_arch_qa} in the main paper summarizes the compact cross-architecture comparison using the best checkpoint for each benchmark, this appendix table reports all intermediate checkpoints across the five self-evolution iterations.
These experiments examine whether the evidence-centered self-evolution procedure transfers beyond the Qwen3-VL backbones used in the main experiments.
Across the two architectures, Video-Zero improves most video QA and reasoning benchmarks, with especially clear gains on reasoning-oriented tasks such as LongVideoReason, VideoMathQA, and VideoMMMU.
The improvements are particularly strong for InternVL3.5-4B on reasoning benchmarks, while long-video understanding shows more backbone- and benchmark-dependent trends.
Overall, the results suggest that evidence-grounded supervision provides transferable training signals across different video VLM backbones.
\begin{table*}[t]
\centering
\caption{Cross-architecture results on long-video understanding and video reasoning benchmarks.
Panel (a) reports results on MiMo-SFT-7B, and panel (b) reports results on InternVL3.5-4B.
Columns v1--v5 denote Video-Zero checkpoints across self-evolution iterations.
Bold numbers indicate the best checkpoint for each benchmark.
$\Delta$ denotes the maximum improvement over the base model.}
\label{tab:cross_arch_qa_side_by_side}
\scriptsize
\setlength{\tabcolsep}{3.0pt}
\renewcommand{\arraystretch}{1.08}

\begin{minipage}[t]{0.485\textwidth}
\centering
\vspace{0pt}
\textbf{(a) MiMo-SFT-7B}
\vspace{0.3em}

\resizebox{\linewidth}{!}{
\begin{tabular}{lccccccc}
\toprule
\textbf{Benchmark} & \textbf{Base} & \textbf{v1} & \textbf{v2} & \textbf{v3} & \textbf{v4} & \textbf{v5} & \textbf{$\Delta$} \\
\midrule
\multicolumn{8}{c}{\textit{Long-video Understanding}} \\
\midrule
LongVideoBench   & 47.60 & 50.00 & \textbf{53.00} & 49.10 & 50.20 & 51.20 & +5.40 \\
MLVU             & 40.40 & 41.90 & 43.00 & 43.60 & 43.30 & \textbf{43.70} & +3.30 \\
Video-MME-Long   & 33.10 & 34.10 & \textbf{35.90} & 34.40 & 34.70 & 34.00 & +2.80 \\
LSDBench         & 51.92 & 54.45 & 54.29 & \textbf{55.83} & 55.67 & 55.14 & +3.91 \\
VideoNIAH        & 46.67 & \textbf{46.81} & 46.74 & 45.26 & 45.41 & 44.96 & +0.14 \\
\midrule
\multicolumn{8}{c}{\textit{Video Reasoning}} \\
\midrule
LongVideoReason  & 69.10 & 72.60 & \textbf{73.10} & 72.50 & 71.80 & 68.50 & +4.00 \\
MMVU             & 69.60 & \textbf{70.08} & \textbf{70.08} & 68.16 & 68.00 & 68.16 & +0.48 \\
VideoMathQA      & 29.76 & 30.48 & \textbf{32.62} & 29.52 & \textbf{32.62} & 27.86 & +2.86 \\
VideoMMMU        & 67.33 & 67.56 & 67.44 & \textbf{68.33} & 67.78 & 67.56 & +1.00 \\
\bottomrule
\end{tabular}
}
\end{minipage}
\hfill
\begin{minipage}[t]{0.485\textwidth}
\centering
\vspace{0pt}
\textbf{(b) InternVL3.5-4B}
\vspace{0.3em}

\resizebox{\linewidth}{!}{
\begin{tabular}{lccccccc}
\toprule
\textbf{Benchmark} & \textbf{Base} & \textbf{v1} & \textbf{v2} & \textbf{v3} & \textbf{v4} & \textbf{v5} & \textbf{$\Delta$} \\
\midrule
\multicolumn{8}{c}{\textit{Long-video Understanding}} \\
\midrule
LongVideoBench   & 60.30 & 61.00 & 60.40 & \textbf{61.20} & 60.70 & 60.90 & +0.90 \\
MLVU             & \textbf{62.40} & 62.00 & 61.80 & 61.80 & 61.90 & 61.50 & +0.00 \\
Video-MME-Long   & 52.40 & 53.30 & 53.30 & 53.30 & 53.20 & \textbf{53.70} & +1.30 \\
LSDBench         & \textbf{60.35} & 57.52 & 57.21 & 57.06 & 56.98 & 57.36 & +0.00 \\
VideoNIAH        & \textbf{38.81} & 36.89 & 36.89 & 36.81 & 36.67 & 36.44 & +0.00 \\
\midrule
\multicolumn{8}{c}{\textit{Video Reasoning}} \\
\midrule
LongVideoReason  & 68.00 & \textbf{72.00} & \textbf{72.00} & 71.40 & 71.60 & 71.70 & +4.00 \\
MMVU             & 54.08 & 57.60 & 58.40 & \textbf{58.88} & 58.72 & 58.40 & +4.80 \\
VideoMathQA      & 12.62 & 17.86 & 21.90 & 23.10 & 23.33 & \textbf{23.81} & +11.19 \\
VideoMMMU        & 44.22 & \textbf{53.11} & 51.78 & 52.56 & 52.00 & 52.56 & +8.89 \\
\bottomrule
\end{tabular}
}
\end{minipage}

\end{table*}

\subsection{Full Ablation Results}
\label{app:full_ablation_results}

The main paper reports ablation results at Iter 4, which serves as a unified comparison point for all ablated variants.
This avoids selecting different best checkpoints for different variants and provides a consistent view of how each component contributes to evidence-centered self-evolution.
Table~\ref{tab:ablation_main} summarizes the category-level averages, where the grounding average is computed over all temporal grounding metrics, and the video QA and reasoning averages are computed over the corresponding benchmark groups used in the main experiments.
The overall average is then computed from these category-level averages.
Tables~\ref{tab:ablation_grounding_iter4} and~\ref{tab:ablation_qa_iter4} provide the detailed Iter 4 results: Table~\ref{tab:ablation_grounding_iter4} reports full temporal grounding metrics, while Table~\ref{tab:ablation_qa_iter4} reports representative long-video understanding and video reasoning benchmarks.

\begin{table*}[t]
\centering
\caption{Full temporal grounding ablation results at Iter 4 on Qwen3-VL-4B.
All variants are evaluated at the same iteration for a unified ablation protocol.}
\label{tab:ablation_grounding_iter4}
\scriptsize
\setlength{\tabcolsep}{3.2pt}
\renewcommand{\arraystretch}{1.08}
\resizebox{\textwidth}{!}{
\begin{tabular}{lcccccccccccccc}
\toprule
\multirow{2}{*}{\textbf{Method}}
& \multicolumn{4}{c}{\textbf{ANet-RTL}}
& \multicolumn{4}{c}{\textbf{ActivityNet}}
& \multicolumn{4}{c}{\textbf{Charades}}
& \multicolumn{2}{c}{\textbf{NExT-GQA}} \\
\cmidrule(lr){2-5}\cmidrule(lr){6-9}\cmidrule(lr){10-13}\cmidrule(lr){14-15}
& \textbf{mIoU} & \textbf{R@0.3} & \textbf{R@0.5} & \textbf{R@0.7}
& \textbf{mIoU} & \textbf{R@0.3} & \textbf{R@0.5} & \textbf{R@0.7}
& \textbf{mIoU} & \textbf{R@0.3} & \textbf{R@0.5} & \textbf{R@0.7}
& \textbf{GQA@0.3} & \textbf{GQA@0.5} \\
\midrule
Video-Zero
& 29.75 & 44.98 & 28.82 & 13.54
& 32.12 & 44.76 & 27.67 & 16.24
& 41.85 & 65.11 & 40.81 & 20.03
& 38.29 & 23.00 \\
\midrule
Description-only Questioner
& 17.70 & 24.45 & 16.59 & 9.61
& 23.59 & 32.18 & 19.78 & 11.72
& 37.85 & 59.57 & 35.97 & 16.91
& 20.94 & 13.56 \\
w/o Video Dependency
& 23.19 & 34.06 & 21.40 & 10.92
& 30.32 & 41.87 & 26.24 & 15.28
& 40.00 & 61.80 & 38.68 & 17.88
& 38.23 & 23.84 \\
w/o Evidence Quality
& 28.24 & 41.48 & 27.07 & 14.41
& 31.28 & 43.66 & 27.22 & 15.75
& 38.38 & 59.41 & 36.21 & 17.10
& 38.59 & 23.61 \\
w/o Learnability
& 22.80 & 31.88 & 21.40 & 9.17
& 29.64 & 41.25 & 24.90 & 14.97
& 39.33 & 60.48 & 37.18 & 18.39
& 36.50 & 22.13 \\
w/o Temporal Reward
& 14.72 & 19.65 & 13.10 & 7.42
& 20.91 & 28.53 & 17.17 & 10.18
& 36.88 & 58.74 & 35.30 & 15.70
& 26.44 & 17.32 \\
Frozen Questioner
& 26.95 & 39.30 & 25.33 & 11.79
& 30.28 & 42.12 & 26.05 & 15.40
& 40.24 & 61.94 & 38.25 & 19.87
& 37.42 & 23.18 \\
\bottomrule
\end{tabular}
}
\end{table*}

\begin{table}[t]
\centering
\caption{Representative long-video understanding and video reasoning ablation results at Iter 4 on Qwen3-VL-4B.
All variants are evaluated at the same iteration; category-level averages in the main paper are computed over the full benchmark groups.}
\label{tab:ablation_qa_iter4}
\small
\setlength{\tabcolsep}{4.8pt}
\renewcommand{\arraystretch}{1.08}
\resizebox{\linewidth}{!}{
\begin{tabular}{lcccccc}
\toprule
\multirow{2}{*}{\textbf{Method}}
& \multicolumn{2}{c}{\textbf{Long-video Understanding}}
& \multicolumn{4}{c}{\textbf{Video Reasoning}} \\
\cmidrule(lr){2-3}\cmidrule(lr){4-7}
& \textbf{Video-MME-L}
& \textbf{LSDBench}
& \textbf{LongVideoReason}
& \textbf{MMVU}
& \textbf{VideoMathQA}
& \textbf{VideoMMMU} \\
\midrule
Video-Zero
& 53.60 & 56.52
& 70.30 & 64.96 & 21.67 & 61.56 \\
\midrule
Description-only Questioner
& 52.70 & 55.29
& 69.40 & 65.60 & 22.14 & 57.89 \\
w/o Video Dependency
& 51.90 & 55.21
& 71.20 & 66.08 & 20.71 & 58.78 \\
w/o Evidence Quality
& 52.00 & 56.98
& 70.90 & 64.96 & 20.24 & 58.78 \\
w/o Learnability
& 52.90 & 54.52
& 70.40 & 64.96 & 21.43 & 60.11 \\
w/o Temporal Reward
& 52.30 & 55.52
& 68.90 & 62.56 & 18.10 & 57.00 \\
Frozen Questioner
& 51.40 & 56.21
& 69.00 & 63.68 & 21.90 & 57.67 \\
\bottomrule
\end{tabular}
}
\end{table}
\subsection{Evidence Analysis Protocol}
\label{app:evidence_analysis}

We provide additional details for the evidence analysis in Figure~\ref{fig:motivation}(a)(b).
The goal is to examine whether questions generated by Video-Zero depend more strongly on temporally localized video evidence than those generated by V-Zero~\citep{wang2026vzero}.
To avoid circular evaluation, we use Qwen3-VL-235B-A22B-Thinking-FP8~\citep{qwen3vl} as an external Oracle model only for annotation, and use a fixed base Solver for all evaluations.

\paragraph{Video dependency.}
Figure~\ref{fig:motivation}(a) evaluates whether generated questions require the video input rather than being answerable from language priors alone.
For each generated question, we evaluate the fixed base Solver under two conditions: with the video and without the video.
We report the with-video accuracy and the video-dependency gap, defined as the difference between with-video and without-video accuracy.
A larger gap indicates that the generated question depends more strongly on visual evidence in the video.
From Iteration 1 to Iteration 4, Video-Zero increases both the with-video accuracy and the dependency gap, while V-Zero shows weaker or decreasing trends.
This suggests that evidence-centered question generation produces supervision that becomes more video-grounded over iterations.

\paragraph{Temporal evidence sensitivity.}
Figure~\ref{fig:motivation}(b) further evaluates whether the generated questions depend on a specific temporal span.
For each generated question, the Oracle first produces a unified reference answer and then annotates the most critical continuous temporal span for answering the question.
All methods and iterations are annotated by the same Oracle, ensuring that comparisons are not affected by method-specific annotation bias.
The Oracle-annotated span is treated as the key temporal evidence for the question.
Given this key span, we construct four controlled frame conditions for Solver evaluation: \textit{Full}, which uses the full video; \textit{Only-key}, which uses only frames inside the key span; \textit{Mask-key}, which removes frames inside the key span and keeps the remaining frames; and \textit{Random}, which uses an equal-length random span that does not overlap with the key span when possible.
The fixed base Solver is evaluated under all four conditions, and its predictions are compared against the Oracle reference answer.

We focus on two diagnostic metrics that quantify dependency on the Oracle-annotated key span:
\begin{equation}
    \mathrm{Key\ Necessity}
    =
    \mathrm{Acc}_{\mathrm{full}}-\mathrm{Acc}_{\mathrm{mask}},
    \qquad
    \mathrm{Key\ Specificity}
    =
    \mathrm{Acc}_{\mathrm{key}}-\mathrm{Acc}_{\mathrm{rand}}.
\end{equation}
Key Necessity measures whether the key span is indispensable by comparing full-video accuracy with accuracy after masking the key span.
Key Specificity measures whether the key span is more informative than an arbitrary span by comparing the key span with an equal-length random span.
Higher values indicate that the generated question depends more strongly on localized temporal evidence rather than generic visual context.

Table~\ref{tab:evidence_analysis_full} reports the Iteration-4 comparison.
Video-Zero achieves a Key Necessity of $27.96\%$ and a Key Specificity of $25.81\%$, compared with $15.05\%$ and $11.83\%$ for V-Zero.
This indicates that Video-Zero generates questions whose answers depend more strongly on specific temporal evidence, supporting the effectiveness of organizing self-generated supervision around localized temporal evidence.

\begin{table}[t]
\centering
\caption{\textbf{Iteration-4 evidence analysis.}
We evaluate a fixed base Solver under controlled frame conditions derived from Oracle-annotated key temporal spans.
Video-Zero shows higher Key Necessity and Key Specificity than V-Zero.}
\label{tab:evidence_analysis_full}
\small
\setlength{\tabcolsep}{6.0pt}
\renewcommand{\arraystretch}{1.10}
\resizebox{0.82\linewidth}{!}{
\begin{tabular}{lcccccc}
\toprule
\textbf{Method} & \textbf{N} &
$\mathbf{Acc_{\rm full}}$ &
$\mathbf{Acc_{\rm key}}$ &
$\mathbf{Acc_{\rm mask}}$ &
$\mathbf{Acc_{\rm rand}}$ &
\textbf{Key Nec.} /
\textbf{Key Spec.} \\
\midrule
Video-Zero & 93 & 64.52 & 54.84 & 36.56 & 29.03 & 27.96 / 25.81 \\
V-Zero     & 93 & 55.91 & 53.76 & 40.86 & 41.94 & 15.05 / 11.83 \\
\bottomrule
\end{tabular}
}
\end{table}

\subsection{Robustness, Scaling, and Qualitative Analysis}
\label{app:robustness_scaling_analysis}

We provide additional analyses to examine the robustness and behavior of Video-Zero beyond the main experimental setting.
Specifically, we evaluate Video-Zero across different frame budgets, training data scales, and random seeds, and further include qualitative examples that illustrate how Questioner-generated supervision evolves over self-evolution iterations.
These analyses complement the main results by testing whether the gains of evidence-centered self-evolution remain stable under changes in visual context length, unlabeled data scale, and training randomness.

\subsubsection{Frame-Budget Robustness}
\label{app:frame_budget_results}

We provide detailed results for the frame-budget robustness analysis in Figure~\ref{fig:frame_budget_grounding}.
All settings use the same 2 FPS sampling rate and differ only in the maximum number of sampled frames.
In addition to the main 32-frame setting, we evaluate Qwen3-VL-4B under 64-frame and 128-frame budgets on temporal grounding benchmarks.
Table~\ref{tab:frame_budget_grounding} reports results for these larger frame budgets.
For compactness, we report the corresponding base model, the Iter 4 checkpoint used for the unified comparison in the main paper, and the peak improvement over the corresponding base model across iterations.

Overall, Video-Zero remains effective under larger frame budgets.
The improvements are especially clear on ANet-RTL, ActivityNet, and NExT-GQA, showing that evidence-centered supervision remains beneficial when more visual context is provided.
The gains across most metrics indicate that Video-Zero is robust beyond the main 32-frame setting, while the mixed Charades trend suggests that temporal evidence selection remains important even with larger frame budgets.

\begin{table}[t]
\centering
\caption{Frame-budget robustness on temporal grounding benchmarks.
We report results under 64-frame and 128-frame max-frame settings on Qwen3-VL-4B, with all settings using 2 FPS sampling.
Iter 4 follows the unified comparison point used in the main paper, and $\Delta_{\mathrm{peak}}$ denotes the maximum improvement over the corresponding base model across iterations.}
\label{tab:frame_budget_grounding}
\small
\setlength{\tabcolsep}{4.5pt}
\renewcommand{\arraystretch}{1.08}
\resizebox{\linewidth}{!}{
\begin{tabular}{llccccc}
\toprule
\textbf{Frames} & \textbf{Checkpoint}
& \textbf{ANet-RTL mIoU}
& \textbf{ActivityNet mIoU}
& \textbf{Charades mIoU}
& \textbf{NExT-GQA@0.3}
& \textbf{NExT-GQA@0.5} \\
\midrule
64f  & Base & 17.89 & 24.41 & 38.77 & 23.36 & 15.52 \\
64f  & Iter 4 & 31.39 & 32.36 & 40.81 & 37.82 & 23.18 \\
64f  & $\Delta_{\mathrm{peak}}$ & +13.50 & +9.94 & +2.04 & +15.12 & +8.12 \\
\midrule
128f & Base & 17.24 & 24.58 & 38.70 & 23.28 & 15.31 \\
128f & Iter 4 & 20.86 & 28.22 & 28.73 & 35.22 & 21.95 \\
128f & $\Delta_{\mathrm{peak}}$ & +6.37 & +5.43 & +0.00 & +12.18 & +6.93 \\
\bottomrule
\end{tabular}
}
\end{table}

\subsubsection{Training Data Scaling}
\label{app:scaling_results}

We study how Video-Zero scales with the amount of unlabeled video data used for self-evolution.
In addition to the 600-video setting used in the main experiments, we evaluate larger unlabeled video pools with 1K and 2K sampled videos while keeping the same Questioner--Solver pipeline, reward design, and optimization procedure.
Tables~\ref{tab:data_scaling_grounding} and~\ref{tab:data_scaling_qa_avg} summarize the peak results across self-evolution iterations for each data scale.

Overall, increasing the unlabeled video pool improves temporal grounding, with the average grounding score increasing from 33.15 at 600 videos to 36.44 at 2K videos.
The gains are especially clear on ANet-RTL, ActivityNet, and Charades, suggesting that a larger raw video pool helps the Questioner discover more diverse temporal evidence.
For long-video understanding and video reasoning, scaling brings additional gains at 2K, although the trend is less monotonic than temporal grounding.
This indicates that additional unlabeled videos mainly strengthen temporal evidence learning, while downstream QA and reasoning improvements also depend on the match between generated supervision and benchmark requirements.

\begin{table}[t]
\centering
\caption{Training data scaling on temporal grounding benchmarks. We report peak results across self-evolution iterations for each unlabeled video scale on Qwen3-VL-4B. Avg. is computed over the five reported metrics.}
\label{tab:data_scaling_grounding}
\small
\setlength{\tabcolsep}{4.5pt}
\renewcommand{\arraystretch}{1.08}
\resizebox{\linewidth}{!}{
\begin{tabular}{lcccccc}
\toprule
\textbf{Data Scale}
& \textbf{ANet-RTL mIoU}
& \textbf{ActivityNet mIoU}
& \textbf{Charades mIoU}
& \textbf{NExT-GQA@0.3}
& \textbf{NExT-GQA@0.5}
& \textbf{Avg.} \\
\midrule
600 & 29.75 & 32.59 & 41.85 & 38.29 & 23.25 & 33.15 \\
1K  & 31.42 & 33.76 & 41.92 & 39.15 & 24.47 & 34.14 \\
2K  & \textbf{32.58} & \textbf{38.21} & \textbf{46.54} & \textbf{40.12} & \textbf{24.76} & \textbf{36.44} \\
\bottomrule
\end{tabular}
}
\end{table}

\begin{table}[t]
\centering
\caption{Training data scaling on long-video understanding and video reasoning. We report peak category averages across self-evolution iterations for each unlabeled video scale on Qwen3-VL-4B.}
\label{tab:data_scaling_qa_avg}
\small
\setlength{\tabcolsep}{7pt}
\renewcommand{\arraystretch}{1.08}
\begin{tabular}{lcc}
\toprule
\textbf{Data Scale}
& \textbf{Long-video Avg.}
& \textbf{Reasoning Avg.} \\
\midrule
600 & 55.88 & 55.31 \\
1K  & 55.62 & 54.59 \\
2K  & \textbf{56.87} & \textbf{55.63} \\
\bottomrule
\end{tabular}
\end{table}

\subsubsection{Random Seed Robustness}
\label{app:multiseed_results}

We evaluate the robustness of Video-Zero under different random seeds.
In addition to the main run, we train two additional runs with different random seeds while keeping the same 600 unlabeled videos, training pipeline, reward design, and evaluation protocol.
Table~\ref{tab:multiseed_category_avg} summarizes the peak iteration-level averages across the three runs.

For temporal grounding, we first average all reported metrics from ANet-RTL, ActivityNet, Charades, and NExT-GQA within each iteration, and then report the best iteration-level average for each seed.
For QA, we average the five long-video understanding benchmarks and four video reasoning benchmarks within each iteration, and then report the best iteration-level QA average for each seed.
This protocol evaluates whether the iterative self-evolution process reliably reaches similar performance under different random seeds, rather than selecting the best checkpoint separately for each benchmark.
Across three seeds, Video-Zero achieves peak iteration-level averages of $32.76 \pm 0.69$ on temporal grounding and $54.48 \pm 0.72$ on QA.
These results show that Video-Zero remains stable across random seeds for both temporal grounding and broader video QA.

\begin{table}[t]
\centering
\caption{Random seed robustness on the Qwen3-VL-4B backbone.
We combine the main run with two additional seeds and report peak iteration-level category averages across iterations.}
\label{tab:multiseed_category_avg}
\footnotesize
\setlength{\tabcolsep}{5.5pt}
\renewcommand{\arraystretch}{1.05}
\resizebox{0.55\linewidth}{!}{
\begin{tabular}{lcc}
\toprule
\textbf{Seed}
& \textbf{Temporal Grounding Avg.}
& \textbf{QA Avg.} \\
\midrule
Main & 33.36 & 55.28 \\
123  & 32.92 & 54.25 \\
456  & 32.00 & 53.90 \\
\midrule
\textbf{Mean $\pm$ Std.}
& \textbf{$32.76 \pm 0.69$}
& \textbf{$54.48 \pm 0.72$} \\
\bottomrule
\end{tabular}
}
\end{table}

\subsubsection{Qualitative Analysis of Questioner-generated Supervision}
\label{app:qualitative_questioner}

\begin{figure*}[t]
    \centering
    \includegraphics[width=0.68\textwidth]{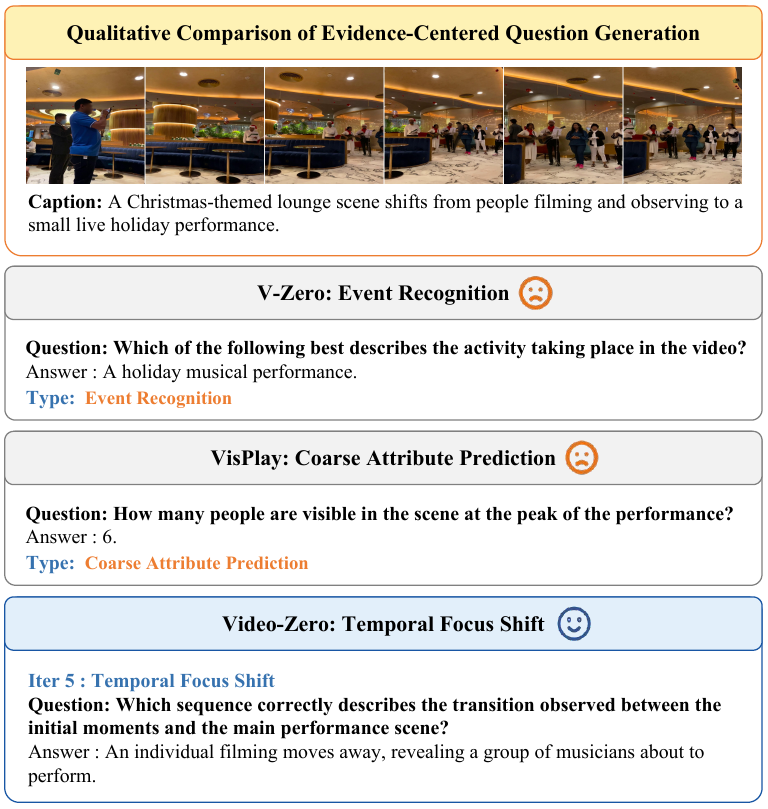}
    \vspace{-0.5em}
    \caption{\textbf{Qualitative comparison on a holiday performance video.}
    Video-Zero generates a temporally grounded question about the transition into the main performance, while V-Zero and VisPlay focus on coarser event or attribute cues.}
    \label{fig:questioner_compare_lounge}
    \vspace{-1.0em}
\end{figure*}

\begin{figure*}[t]
    \centering
    \includegraphics[width=0.68\textwidth]{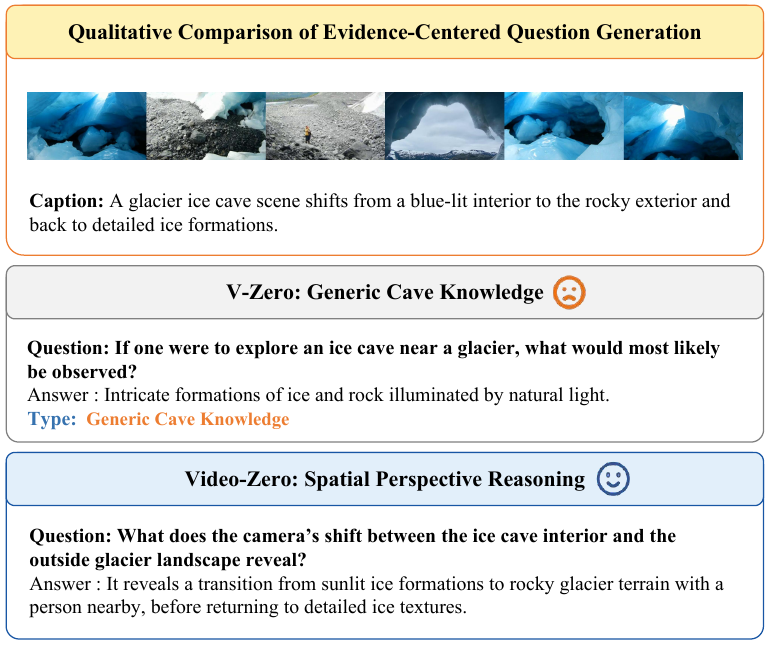}
    \vspace{-0.5em}
    \caption{\textbf{Qualitative comparison on a glacier ice-cave video.}
    Video-Zero captures the camera's spatial perspective shift across cave interior, glacier exterior, and ice formations, while V-Zero asks a generic commonsense question.}
    \label{fig:questioner_compare_icecave}
    \vspace{-1.0em}
\end{figure*}

We provide qualitative case studies from two complementary perspectives.
Figures~\ref{fig:questioner_compare_lounge} and~\ref{fig:questioner_compare_icecave} compare questions generated by different self-evolution methods on the same videos.
V-Zero tends to produce coarse event-level or commonsense questions, while VisPlay often focuses on surface-level attributes or fails to produce a valid formatted question in some cases.
By contrast, Video-Zero generates questions that are more tightly coupled with temporal evidence, visual transitions, and fine-grained scene changes.

\begin{figure*}[t]
    \centering
    \includegraphics[width=0.95\textwidth]{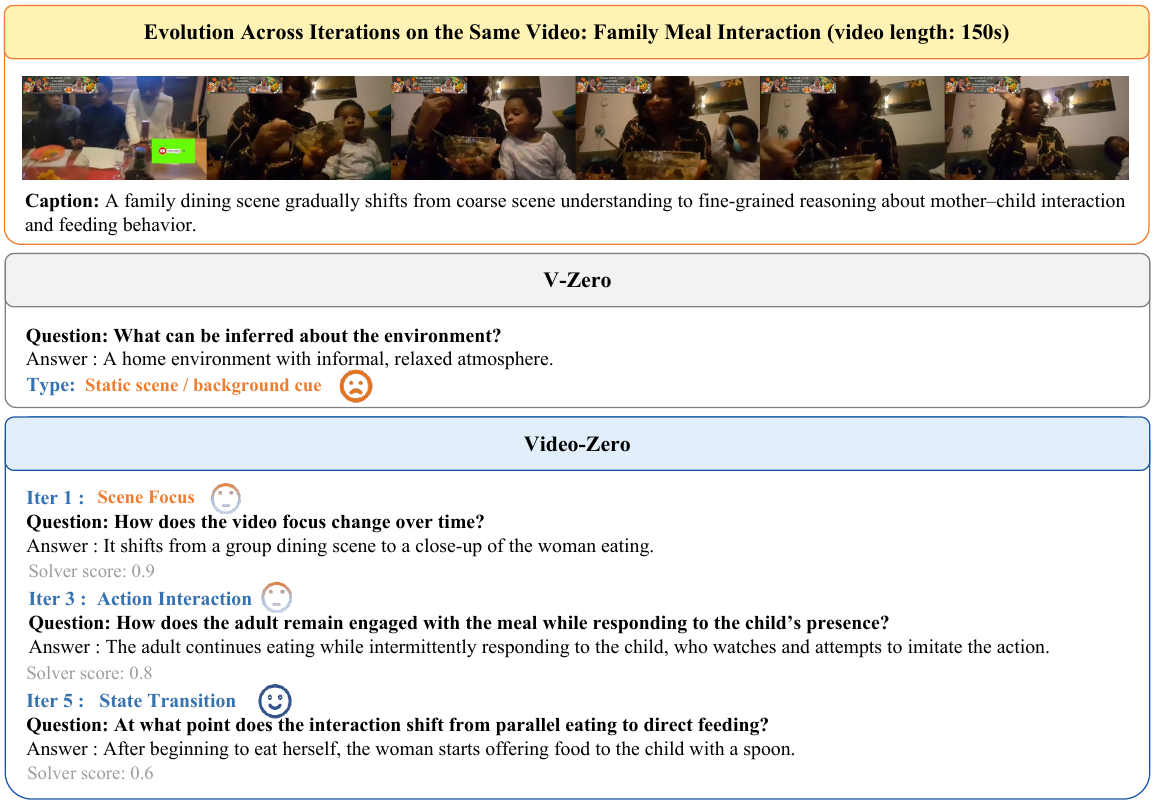}
    \vspace{-0.5em}
    \caption{\textbf{Evolution of Video-Zero Questioner on a family meal video.}
    Across iterations, the questions evolve from coarse scene focus to fine-grained reasoning about mother--child interaction and direct feeding.}
    \label{fig:questioner_evolve_family}
    \vspace{-1.0em}
\end{figure*}

\begin{figure*}[t]
    \centering
    \includegraphics[width=0.95\textwidth]{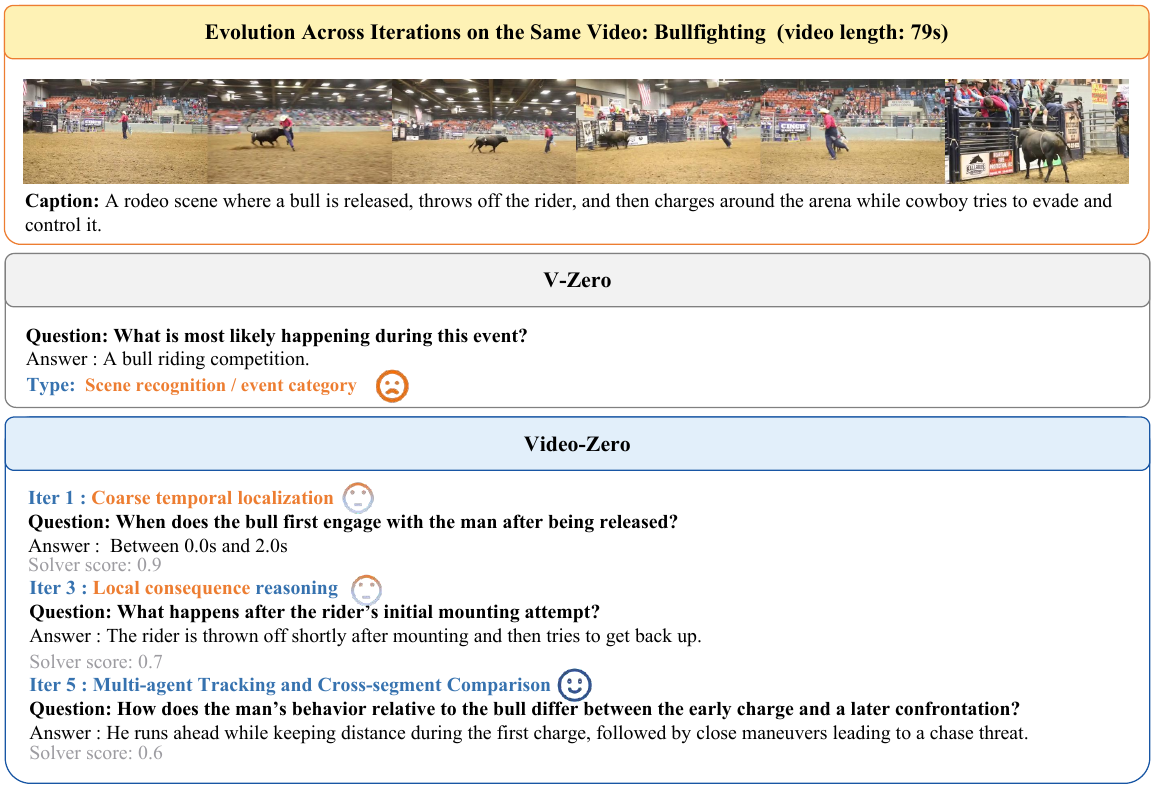}
    \vspace{-0.5em}
    \caption{\textbf{Evolution of Video-Zero Questioner on a bullfighting video.}
    Across iterations, the questions progress from coarse temporal localization to consequence reasoning and multi-agent tracking.}
    \label{fig:questioner_evolve_bullfighting}
    \vspace{-1.0em}
\end{figure*}

Figures~\ref{fig:questioner_evolve_family} and~\ref{fig:questioner_evolve_bullfighting} further illustrate how the Video-Zero Questioner evolves across self-evolution iterations on the same video.
Early iterations tend to produce coarse scene-level questions or short-horizon temporal queries, whereas later iterations generate more fine-grained, interaction-aware, and temporally structured supervision.
These examples suggest that evidence-centered self-evolution improves not only final performance, but also the video-dependence and temporal specificity of the generated supervision.

\subsection{Limitations, Broader Impacts, and Future Work}
\label{app:limitations}

Video-Zero demonstrates that video VLMs can self-evolve from unlabeled videos by organizing generated supervision around temporally localized evidence.
Our results across multiple model backbones show that Video-Zero consistently improves temporal grounding, long-video understanding, and video reasoning performance.
A positive impact of this work is that it reduces the need for manually annotated video QA and temporal grounding data, potentially making video model development more scalable and accessible.

However, the primary limitation of this work is that we were unable to validate this scaling trend on larger video VLMs, such as 32B, 72B, or 100B+ models, due to the prohibitive computational cost.
This leaves the full scaling behavior of evidence-centered self-evolving training as an important direction for future work.
Another limitation is that Video-Zero relies on model-generated questions, answers, and temporal evidence, which may inherit biases or errors from the underlying video VLMs.
For sensitive video domains, additional caution and domain-specific validation would be important.
Future work will explore larger backbones, stronger filtering and verification, and responsible release with documentation, intended-use guidance, and misuse restrictions.

\end{document}